\pdfoutput=1
\documentclass[10pt, logo, copyright]{nv}

\usepackage{graphicx}
\definecolor{nvidiagreen}{HTML}{76B900}

\usepackage{mdframed}
\usepackage{color}
\usepackage{xcolor}
\usepackage[utf8]{inputenc} 
\usepackage[T1]{fontenc}    

\usepackage{amsfonts}       
\usepackage{nicefrac}       
\usepackage{microtype}      
\usepackage{multirow}
\usepackage{multicol}
\usepackage{tabto}
\usepackage{xspace}
\usepackage{amsmath}
\usepackage{adjustbox}
\usepackage{enumitem}
\usepackage{wrapfig}
\usepackage{dblfloatfix}
\usepackage{times}
\usepackage{verbatim}
\usepackage{amssymb}
\usepackage{mathtools}
\usepackage{caption}
\usepackage{subcaption}
\usepackage{array}
\usepackage{colortbl}
\usepackage{booktabs}
\usepackage{bbm}
\usepackage{makecell}
\usepackage{float}
\usepackage{siunitx}
\usepackage{pifont}
\usepackage{marvosym}
\usepackage{listings}
\usepackage{pdflscape}
\usepackage{url}            
\usepackage{booktabs}       
\usepackage{amsfonts}       
\usepackage{nicefrac}       
\usepackage{microtype}      
\usepackage{xcolor}         
\usepackage{tabularx}
\usepackage{amsmath, amssymb}
\usepackage{algorithm}
\usepackage{algpseudocode}
\usepackage{mathrsfs}
\usepackage{natbib}
\usepackage{booktabs}
\usepackage{colortbl}
\usepackage{multirow}
\usepackage{mathtools}
\usepackage{wrapfig} 
\usepackage{subcaption}
\usepackage{footmisc}
\definecolor{codebg}{RGB}{245, 245, 245} 
\definecolor{keywordcolor}{RGB}{0, 0, 153} 
\definecolor{commentcolor}{RGB}{34, 139, 34} 
\definecolor{stringcolor}{RGB}{163, 21, 21}
\definecolor{numbercolor}{RGB}{128, 128, 128}
\usepackage{url}
\usepackage{tabularx}
\usepackage{arydshln}
\usepackage{hhline}
\usepackage{diagbox}
\usepackage{tcolorbox}
\usepackage[nameinlink]{cleveref}
\usepackage{fp} 
\usepackage{authblk}  
\crefname{equation}{Eq.}{Eqs.}
\crefname{figure}{Fig.}{Figs.}
\crefname{algorithm}{Algo}{Algo}
\Crefname{thm}{Thm}{Thm}
\newtheorem{theorem}{Theorem}
\newtheorem{remark}{Remark}

\newcommand{\KLDiv}[2]{D_{\mathrm{KL}}\left(#1 \| #2\right)}
\newcommand{\LpDist}[3]{D_{#1}\left(#2, #3\right)}
\newcommand{\argmax}{\operatornamewithlimits{argmax}}
\newcommand{\calV}{\mathcal{V}}
\newcommand{\bfX}{\boldsymbol{X}}
\newcommand{\bfx}{\boldsymbol{x}}
\newcommand{\bfz}{\boldsymbol{z}}

\usepackage{xcolor}
\definecolor{pearDark}{RGB}{34,139,34}  
\definecolor{mygreen}{RGB}{34,139,34}
\definecolor{mylightblue}{RGB}{0,162,230}
\definecolor{deepyellow}{RGB}{255,215,0}
\definecolor{nvgreen}{RGB}{118, 185, 0}

\usepackage{xspace}

\newcommand{\method}{Fast-dLLM\xspace}

\title{\method: Training-free Acceleration of  Diffusion LLM by Enabling KV Cache and Parallel Decoding}

\author{
Chengyue Wu\textsuperscript{1,2$*$} ~~
Hao Zhang\textsuperscript{2$*$} ~~
Shuchen Xue\textsuperscript{4} ~~
Zhijian Liu\textsuperscript{2} ~~
Shizhe Diao\textsuperscript{2} ~~
Ligeng Zhu\textsuperscript{2} ~~\quad \quad \quad \quad \quad \quad \quad \quad
Ping Luo\textsuperscript{1} ~~
Song Han\textsuperscript{2,3} ~~
Enze Xie\textsuperscript{2} \\
\vspace{2mm}
{\normalsize \textsuperscript{1}The University of Hong Kong ~~
\textsuperscript{2}NVIDIA ~~
\textsuperscript{3}MIT ~~
\textsuperscript{4}Independent Researcher ~~
}\\
{\footnotesize $^*$Equal contribution.}
}

\begin{abstract}
\textbf{Abstract:} Diffusion-based large language models (Diffusion LLMs) have shown promise for non-autoregressive text generation. However, the practical inference speed of open-sourced Diffusion LLMs often lags behind autoregressive models due to the lack of Key-Value (KV) Cache and quality degradation when decoding multiple tokens simultaneously. To bridge this gap, we introduce \method, a method that incorporates a novel block-wise approximate KV Cache mechanism tailored for bidirectional diffusion models, enabling cache reuse with negligible performance drop. Additionally, we identify the root cause of generation quality degradation in parallel decoding as the disruption of token dependencies under the conditional independence assumption. To address this, \method also proposes a confidence-aware parallel decoding strategy that selectively decodes tokens exceeding a confidence threshold, mitigating dependency violations and maintaining generation quality. Experimental results on LLaDA and Dream models across multiple LLM benchmarks demonstrate up to 27.6× throughput improvement with minimal accuracy loss, closing the performance gap with autoregressive models and paving the way for practical deployment of Diffusion LLMs.
    \newline
    \textbf{Links:} \hspace{2pt}
    {\hypersetup{urlcolor=nvidiagreen}
    \href{https://github.com/NVlabs/Fast-dLLM}{Github Code} |
    \href{https://nvlabs.github.io/Fast-dLLM} {Project Page}
    }
\end{abstract}

\begin{document}

\maketitle

\begin{abstract}
Diffusion-based large language models (Diffusion LLMs) have shown promise for non-autoregressive text generation with parallel decoding capabilities. However, the practical inference speed of open-sourced Diffusion LLMs often lags behind autoregressive models due to the lack of Key-Value (KV) Cache and quality degradation when decoding multiple tokens simultaneously. To bridge this gap, we introduce a novel block-wise approximate KV Cache mechanism tailored for bidirectional diffusion models, enabling cache reuse with negligible performance drop. Additionally, we identify the root cause of generation quality degradation in parallel decoding as the disruption of token dependencies under the conditional independence assumption. To address this, we propose a confidence-aware parallel decoding strategy that selectively decodes tokens exceeding a confidence threshold, mitigating dependency violations and maintaining generation quality. Experimental results on LLaDA and Dream models across multiple LLM benchmarks demonstrate up to \textbf{27.6$\times$ throughput} improvement with minimal accuracy loss, closing the performance gap with autoregressive models and paving the way for practical deployment of Diffusion LLMs.
\end{abstract}

\section{Introduction}
\begin{figure*}[ht]
    \centering
    \begin{subfigure}[t]{0.48\textwidth}
        \centering
        \includegraphics[width=\textwidth]{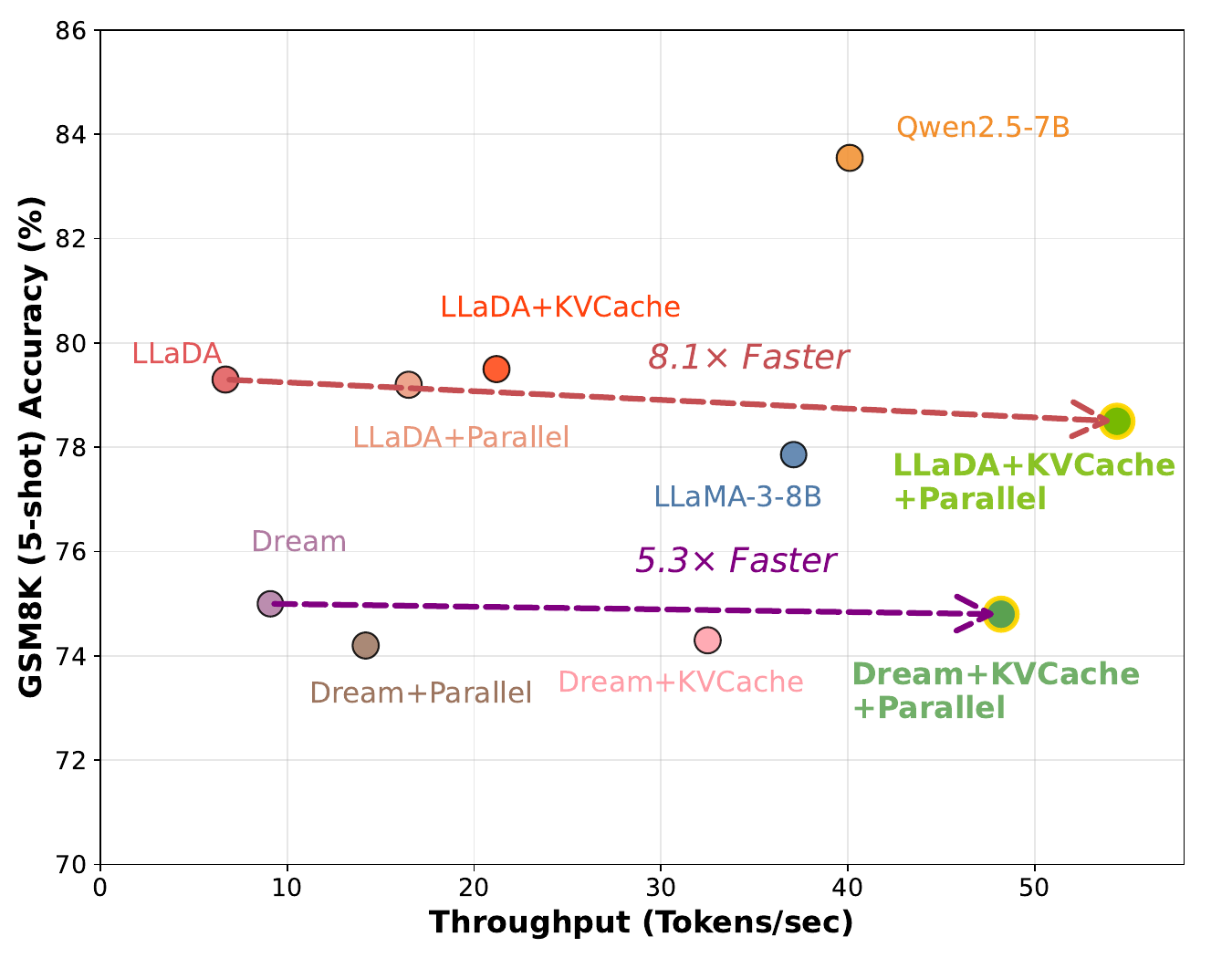}
        \caption{Throughput vs. Accuracy across methods}
    \end{subfigure}
    \hfill
    \begin{subfigure}[t]{0.48\textwidth}
        \centering
        \includegraphics[width=\textwidth]{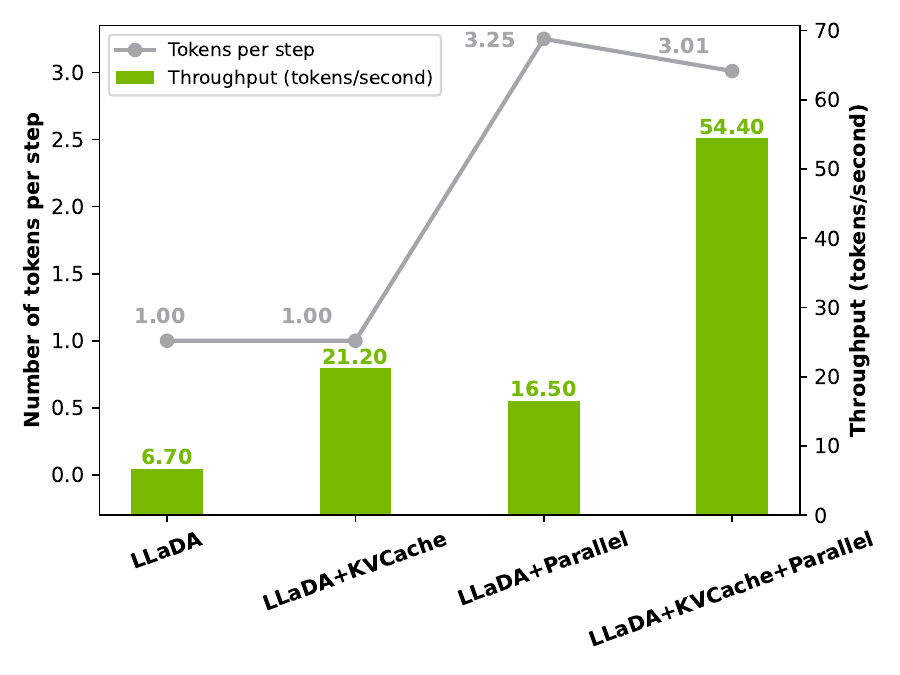}
        \caption{Throughput and tokens per step across methods}
    \end{subfigure}

    \begin{subfigure}[t]{\textwidth}
        \centering
        \includegraphics[width=\textwidth]{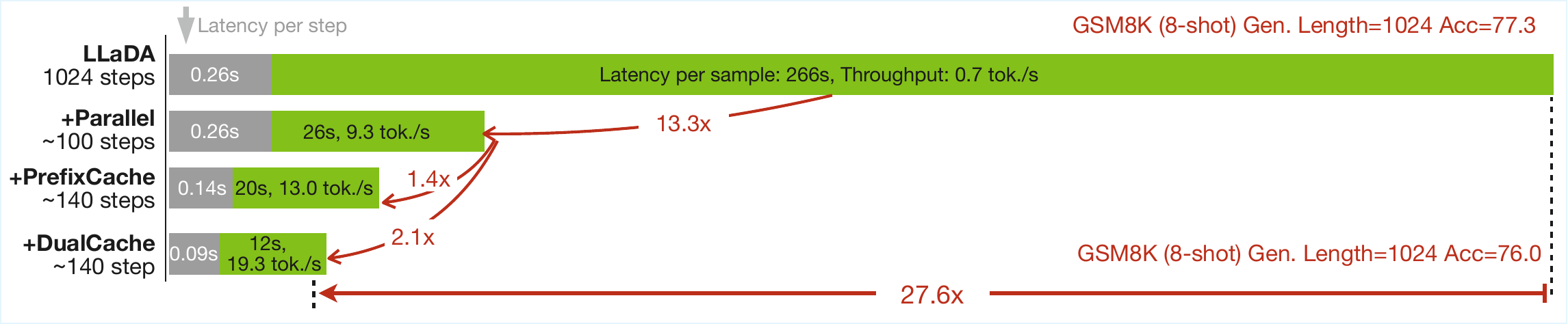}
        \caption{End-to-end speedup over vanilla LLaDA baseline.}
    \end{subfigure}

    \caption{
        \textbf{Effectiveness of components of \method across different approaches.} We use NVIDIA A100 GPU  with a single batch size and no inference speedup frameworks..
        \textbf{(a)} Inference throughput (tokens/sec) and GSM8K (5-shot) accuracy across various designs and models under a maximum generation length of 256. Caching mechanism and parallel decoding can significantly accelerate inference, while the combination provides up to an 8.1$\times$ increase in throughput with negligible accuracy reduction. 
        \textbf{(b)} We break down the contributions of each method by showing both the number of tokens generated per step (line) and total throughput (bars). 
        \textbf{(c)} With long prefilling (8-shot) and a maximum generation length of 1024, our combined approach achieves up to 27.6$\times$ end-to-end speedup compared to the vanilla LLaDA baseline.
    }
    \label{fig:teaser}
\end{figure*}

Diffusion-based large language models (Diffusion LLMs) have recently attracted increasing attention due to their potential for parallel token generation and the advantages of bidirectional attention mechanisms. Notably, Mercury~\cite{mercury2025} runs at over 1,000 tokens per second, and Gemini Diffusion~\cite{gemini_diffusion2025} by Google DeepMind has demonstrated the ability to generate over 1,400 tokens per second, highlighting the promise of significant inference acceleration.

However, current open-source Diffusion LLMs~\cite{nie2025largelanguagediffusionmodels,dream2025} have yet to close such throughput gap in practice, and their actual speed often falls short of autoregressive (AR) models. This is primarily due to two issues. First, diffusion LLMs do not support key-value (KV) caching, a critical component in AR models for speeding up inference. Second, the generation quality tends to degrade when decoding multiple tokens in parallel. For example, recent findings such as those from LLaDA~\cite{nie2025largelanguagediffusionmodels} indicate that Diffusion LLMs perform best when generating tokens one at a time and soon degrades when decoding multiple tokens simultaneously.

To bridge the performance gap with AR models that benefit from KV Cache, we present \method, a fast and practical diffusion-based language modeling framework. First, \method introduces an approximate KV Cache tailored to Diffusion LLMs. While the bidirectional nature of attention in Diffusion LLMs precludes a fully equivalent KV Cache, our approximation closely resembles an ideal cache in practice. To support KV Cache, we adopt a block-wise generation manner. Before generating a block, we compute and store KV Cache of the other blocks to reuse. After generating the block, we recompute the KV Cache of all the blocks. Visualizations confirm the high similarity with adjacent inference steps within the block, and our experiments show that this approximation preserves model performance during inference. We further propose a DualCache version that caches Keys and Values for both prefix and suffix tokens.

In parallel, \method investigates the degradation in output quality when generating multiple tokens simultaneously. Through theoretical analysis and empirical studies, we identify that simultaneous sampling of interdependent tokens under a conditional independence assumption disrupts critical token dependencies. To address this issue and fully exploit the parallelism potential of Diffusion LLMs, we propose a novel confidence-thresholding strategy to select which tokens can be safely decoded simultaneously. Instead of selecting the tokens with top K confidence to decode as in LLaDA, we select tokens with confidence larger than a threshold. Our theoretical justification and experimental results demonstrate that this strategy maintains generation quality while achieving up to 13.3$\times$ inference speed-up.

In summary, our contributions are threefold:

\begin{enumerate}
\item \textbf{Key-Value Cache for Block-Wise Decoding} We introduce a block-wise approximate KV Cache mechanism specifically designed for bidirectional attention. Our approach reuses cached activations from previously decoded blocks by exploiting the high similarity of KV activations between adjacent steps. By caching both prefix and suffix blocks, the DualCache strategy enables substantial computational reuse.
\item \textbf{Confidence-Aware Parallel Decoding} We propose a novel confidence-aware parallel decoding method. Unlike prior approaches that select a fixed number of tokens per step, our method dynamically selects tokens whose confidence exceeds a global threshold, enabling safe and effective parallel decoding. This approach significantly accelerates inference by 13.3$\times$ while preserving output quality.
\item \textbf{State-of-the-Art Acceleration Results} We conduct comprehensive experiments on multiple open-source Diffusion LLMs (LLaDA, Dream) and four mainstream benchmarks (GSM8K, MATH, HumanEval, MBPP). Results demonstrate that our \method consistently deliver order-of-magnitude speedups with minimal or no degradation in accuracy, confirming the generality and practical value of our approach for real-world deployment. \method achieves hgiher acceleration (up to 27.6$\times$) when generation length is longer ($1024$).
\end{enumerate}
\section{Preliminary}
\subsection{Masked Diffusion Model}

Diffusion models for discrete data were first explored in~\cite{sohl2015deep,hoogeboom2021argmax}.
Subsequently, D3PM~\cite{austin2021structured} proposed a more general framework, defining the forward noising process via a discrete state Markov chain with specific transition matrices $\boldsymbol{Q}_t$, and parameterized $p_{\theta}(\boldsymbol{x}_0 | \boldsymbol{x}_t)$ for learning the reverse process by maximizing the Evidence Lower Bound (ELBO).
CTMC~\cite{campbell2022continuous} further extended D3PM to continuous time, formalizing it within a continuous-time Markov Chain (CTMC) framework.
In a different approach, SEDD~\cite{lou2023discrete} parameterizes the likelihood ratio $\frac{p_t(\boldsymbol{y})}{p_t(\boldsymbol{x})}$ for learning the reverse process, and employs Denoising Score Entropy to train this ratio.

Among the various noise processes in discrete diffusion, Masked Diffusion Models (MDMs), also termed absorbing state discrete diffusion models, have gained considerable attention. MDMs employ a forward noising process where tokens are progressively replaced by a special $[\text{MASK}]$ token. This process is defined by the transition probability:
\begin{equation}\label{eq:forward mask}
q_{t|0}\left(\boldsymbol{x}_t | \boldsymbol{x}_0\right) = \prod_{i=1}^n q_{t|0}\left(\boldsymbol{x}_t^i | \boldsymbol{x}_0^i\right) = \prod_{i=1}^n \text{Cat}\left(\boldsymbol{x}_t^i;(1-t)\delta_{\boldsymbol{x}_0^i} + t \delta_{[\text{MASK}]}\right).
\end{equation}
Here, $t \in [0, 1]$ denotes the diffusion time (or masking level), controlling the interpolation between the original data $\boldsymbol{x}_0$ (at $t=0$) and a fully masked sequence (at $t=1$).

More recently, work by MDLM~\cite{shi2024simplified,sahoo2024simple,zheng2024masked} and RADD~\cite{ou2024your} has shown that for MDMs, different parameterizations are equivalent.
Furthermore, they demonstrated that the training objective for MDMs can be simplified or directly derived from the data likelihood. This leads to the following objective function, an Evidence Lower Bound (ELBO) on $\log p_{\boldsymbol{\theta}}(\boldsymbol{x})$:
\begin{equation}
-\log p_{\boldsymbol{\theta}}\left(x\right) \leq \int_{0}^1 \frac{1}{t} \mathbb{E}_{q_{t|0}\left(\boldsymbol{x}_t | \boldsymbol{x}_0\right)}\left[ \sum_{i:\boldsymbol{x}_0^i = [\text{MASK}]} -\log p_{\boldsymbol{\theta}}(\boldsymbol{x}_0^i|\boldsymbol{x}_t)\right] \mathrm{d}t \coloneqq \mathcal{L}_{\text{MDM}}.
\end{equation}

\subsection{Generation Process of MDMs}
The analytical reverse of the forward process defined in Equation~\ref{eq:forward mask} is computationally inefficient for generation, as it typically involves modifying only one token per step~\cite{campbell2022continuous, lou2023discrete}.
A common strategy to accelerate this is to employ a $\tau$-leaping~\cite{gillespie2001approximate} approximation for the reverse process.
In the context of MDMs, this allows for an iterative generation process where multiple masked tokens can be approximately recovered in a single step from a noise level $t$ to an earlier level $s < t$.
\begin{align}
q_{s|t} = \prod_{i=0}^{n-1} q_{s|t}(\boldsymbol{x}_s^i|\boldsymbol{x}_t),\text{ where } 
q_{s|t}(\boldsymbol{x}_s^i|\boldsymbol{x}_t)=\begin{cases}
1, & \boldsymbol{x}_t^i \neq [\text{MASK}], \boldsymbol{x}_s^i = \boldsymbol{x}_t^i \\
\frac{s}{t}, & \boldsymbol{x}_t^i = [\text{MASK}], \boldsymbol{x}_s^i = [\text{MASK}] \\
\frac{t-s}{t} q_{0|t}(\boldsymbol{x}_s^i | \boldsymbol{x}_t), & \boldsymbol{x}_t^i = [\text{MASK}], \boldsymbol{x}_s^i \neq [\text{MASK}].
\end{cases}
\label{eq:q_st_definition_main}
\end{align}
Here, $q_{0|t}(\boldsymbol{x}_s^i | \boldsymbol{x}_t)$ (when $\boldsymbol{x}_t^i = [\text{MASK}]$) represents a distribution over the vocabulary for predicting a non-$[\text{MASK}]$ token, provided by the model. In scenarios involving conditional data, such as generating a response $\boldsymbol{x}_0$ to a prompt $p$, the MDM's reverse process, as defined in Equation~\ref{eq:q_st_definition_main}, requires adaptation.
Specifically, the model's predictive distribution $q_{0|t}(\boldsymbol{x}_s^i | \boldsymbol{x}_t)$ for unmasking a token $\boldsymbol{x}_s^i$ is now also conditioned on the prompt $p$, as $q_{0|t}(\boldsymbol{x}_s^i | \boldsymbol{x}_t, p)$.

\paragraph{Curse of Parallel Decoding} Directly reversing the forward process from Equation~\ref{eq:forward mask} for generation is slow, typically altering just one token per step~\cite{campbell2022continuous, lou2023discrete}. A common strategy to accelerate this is to employ a $\tau$-leaping~\cite{gillespie2001approximate} approximation for the reverse process.
For MDMs, this means multiple masked tokens will be generated in parallel in a single step. However, a significant challenge arises in multiple token prediction due to the conditional independence assumption. Consider an example from~\cite{song2025ideas}: \textit{The list of poker hands that consist of two English words are: $\_\ \_$}. The subsequent two words could be, for instance, ``high card,'' ``two pair,'' ``full house,'' or ``straight flush.'' Notably, a correlation exists between these two words. However, the multi-token prediction procedure in MDMs first generates a probability distribution for each token and then samples from these distributions independently. This independent sampling can lead to undesirable combinations, such as ``high house.''

To formalize this, consider unmasking two token positions, $i$ and $j$. MDMs sample these from $p(\boldsymbol{x}_s^i | \boldsymbol{x}_t) \cdot p(\boldsymbol{x}_s^j | \boldsymbol{x}_t)$ due to the conditional independence assumption. However, the true joint probability requires accounting for the dependency: $p(\boldsymbol{x}_s^i, \boldsymbol{x}_s^j | \boldsymbol{x}_t) = p(\boldsymbol{x}_s^i | \boldsymbol{x}_t) \cdot p(\boldsymbol{x}_s^j | \boldsymbol{x}_t, \boldsymbol{x}_s^i)$ (or symmetrically, by conditioning $i$ on $j$). This discrepancy between the assumed independent generation and the true dependent data distribution can degrade the quality and coherence of the generated sequences.
The issue is more problematic when a large number of tokens are unmasked simultaneously in a single step.

\section{Methodology}
\subsection{Pipeline Overview}

Our approach, \method, builds on the Masked Diffusion Model (MDM) architecture to enable efficient and high-quality sequence generation. To accelerate inference, the overall pipeline incorporates two key strategies: efficient attention computation through Key-Value (KV) Cache and a parallel decoding scheme guided by prediction confidence.

Specifically, we adopt Key-Value Cache for Block-Wise Decoding, which allows reusing attention activations across steps and significantly reduces redundant computation. Within each block, we further propose Confidence-Aware Parallel Decoding, enabling selective updates of tokens based on confidence scores to improve efficiency while maintaining output quality.

By combining these strategies, \method significantly speeds up inference for MDMs with minimal impact on generation performance. The overall procedure is summarized in Algorithm~\ref{alg:pipeline}.
\subsection{Key-Value Cache for Block-Wise Decoding}
\begin{figure*}
    \centering
    \includegraphics[width=\linewidth]{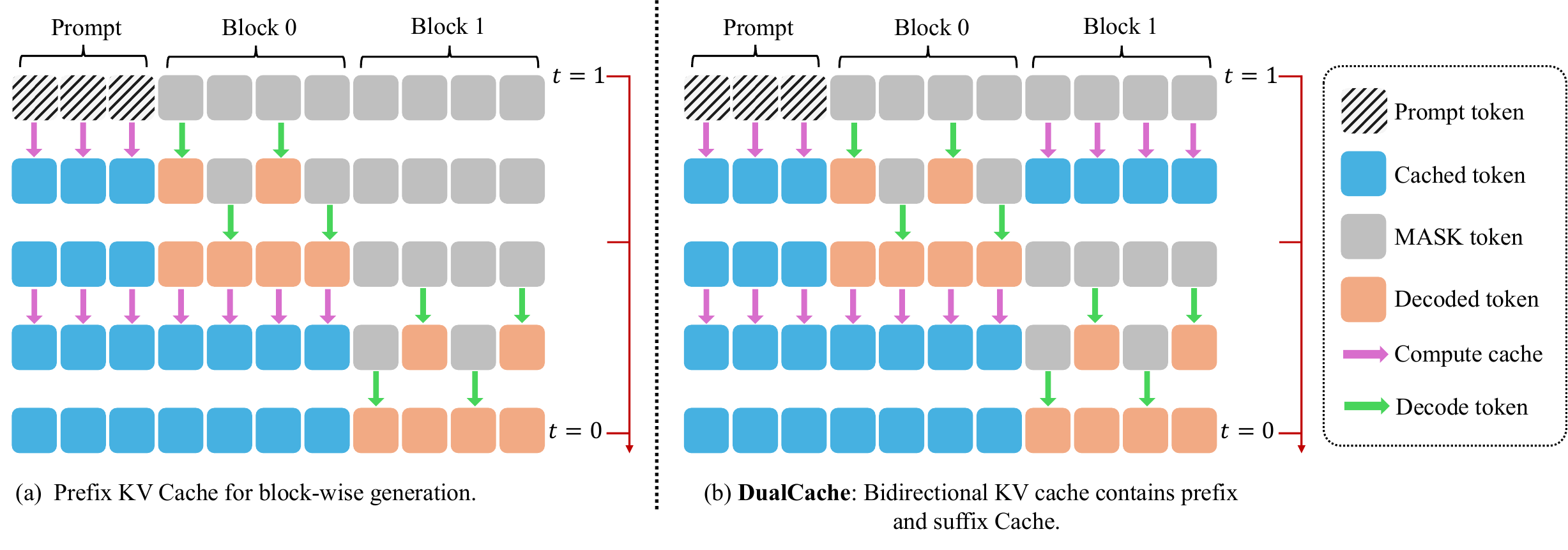}
    \caption{\textbf{Illustration of our Key-Value Cache for Block-Wise Decoding.} (a) During prefix-only caching, the KV cache is computed once for the prompt and reused across multiple decoding steps within each block. The cache is updated after completing a block to maintain consistency, with negligible overhead. 
    (b) DualCache extends this approach by caching both prefix and masked suffix tokens, further accelerating decoding. The high similarity of KV activations across steps allows effective reuse with minimal approximation error.
    }
    \label{fig: kv-cache}
\end{figure*}
As shown in Figure \ref{fig: kv-cache}, we adopt a block-wise decoding strategy to support the use of a Key-Value (KV) Cache. Initially, we compute and store the KV Cache for the prompt, which is reused throughout Block $0$. Within each block, the same cache is reused for multiple decoding steps. After completing the decoding of a block, we update the cache for all tokens (not just the newly generated ones). This cache update can be performed jointly with the decoding step, so compared to not using caching, there is no additional computational overhead. This approach results in an approximate decoding process, due to the use of full attention in masked diffusion models~\cite{nie2025largelanguagediffusionmodels,dream2025}.

\begin{figure*}[htbp]
    \centering
    \begin{subfigure}[t]{0.48\linewidth}
        \centering
        \includegraphics[width=\linewidth]{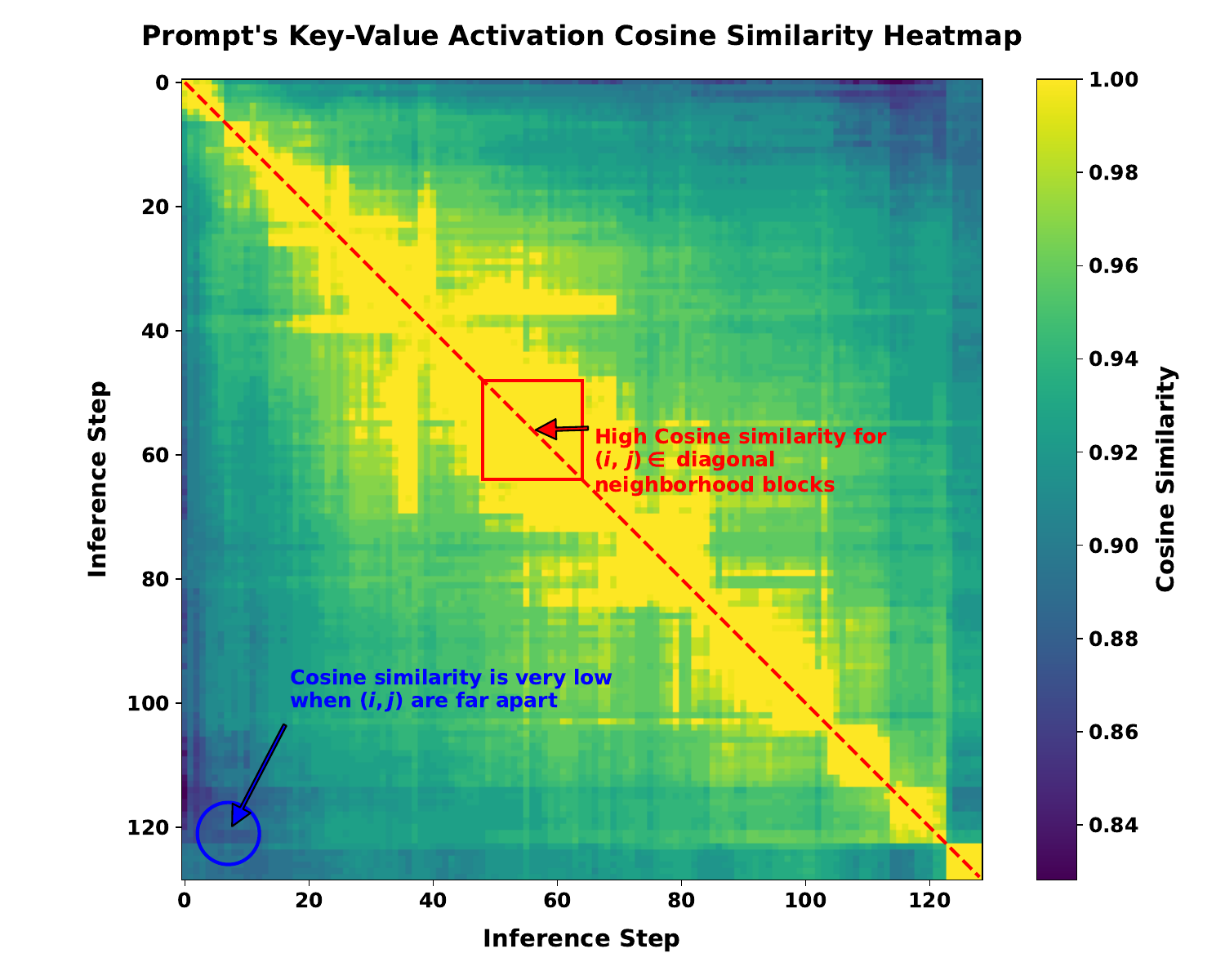}
        \caption{Prompt block}
        \label{fig:prompt-kv-similarity}
    \end{subfigure}
    \hfill
    \begin{subfigure}[t]{0.48\linewidth}
        \centering
        \includegraphics[width=\linewidth]{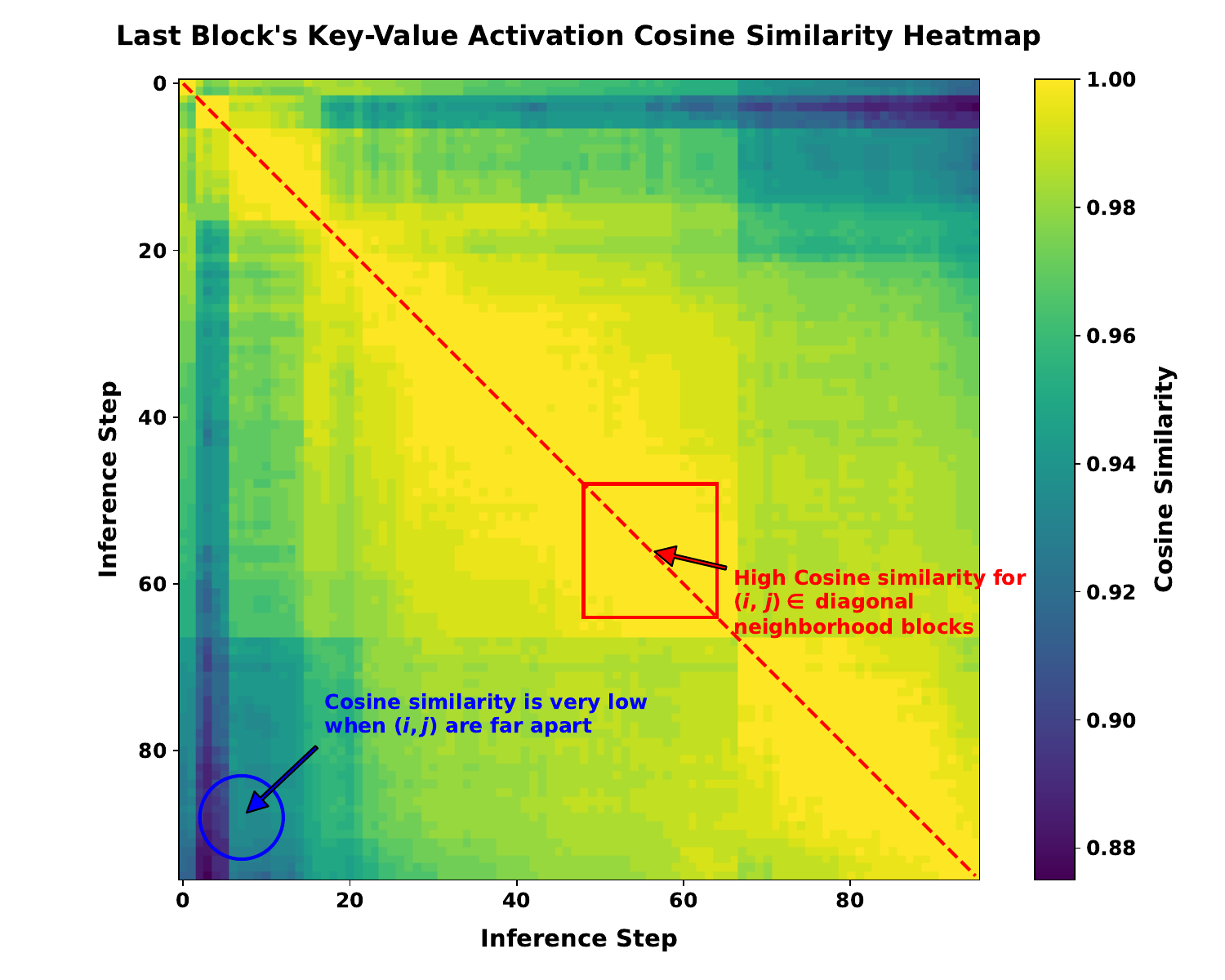}
        \caption{Last block}
        \label{fig:last-block-kv-similarity}
    \end{subfigure}
    \caption{
    \textbf{Heatmaps of Key-Value Activation Cosine Similarity Across Inference Steps in LLaDA-Instruct.}
(a) Cosine similarity heatmap for the prompt block, averaged over all prompt tokens.
(b) Cosine similarity heatmap for the last block, averaged over all tokens in the last block (used to represent suffix tokens, as the last block always belongs to the suffix before its own decoding).
In both (a) and (b), high similarity is observed near the diagonal ($i \approx j$), indicating that Key-Value activations at adjacent inference steps within a block are highly similar. The red boxed regions highlight this effect, supporting the use of an approximate block-wise KV Cache: cached activations from previous steps can be safely reused during block decoding with minimal loss in accuracy. The DualCache strategy, which additionally caches suffix tokens, further demonstrates negligible differences in activations during block decoding, enabling greater acceleration with competitive accuracy.
    }
    \label{fig:kv-cosine-similarity-comparison}
\end{figure*}

The effectiveness of our approximate KV Cache approach stems from the observation that KV activations exhibit high similarity across adjacent inference steps, as illustrated in Figure~\ref{fig:kv-cosine-similarity-comparison}. The red boxed region in Figure~\ref{fig:prompt-kv-similarity} highlights the similarity scores within a block, which are consistently close to 1. This indicates that the differences in prefix keys and values during block decoding are negligible, allowing us to safely reuse the cache without significant loss in accuracy.

Furthermore, we implement a bidirectional version of our KV caching mechanism, named DualCache, that caches not only the prefix tokens but also the suffix tokens, which consist entirely of masked tokens under our block-wise decoding scheme. As shown in Table~\ref{tab:5v8shot}, DualCache results in further acceleration. The red boxed region in Figure~\ref{fig:last-block-kv-similarity} further demonstrates that the differences in suffix keys and values during block decoding are negligible.

\subsection{Confidence-Aware Parallel Decoding}
While approaches like employing auxiliary models to explicitly capture these dependencies exist~\cite{liu2024discrete,xu2024energy}, they typically increase the complexity of the overall pipeline. In contrast to these approaches, we propose a simple yet effective confidence-aware decoding algorithm designed to mitigate this conditional independence issue.

Concretely, at each iteration, rather than aggressively unmasking all masked tokens using their independent marginal probabilities, we compute a confidence score for each token (e.g., the maximum softmax probability). Only those with confidence exceeding a threshold are unmasked in the current step; the rest remain masked and are reconsidered in future steps. If no token’s confidence exceeds the threshold, we always unmask the token with the highest confidence to ensure progress and prevent an infinite loop. This strategy accelerates generation while reducing errors from uncertain or ambiguous predictions.

A critical question, however, is: \textit{When is it theoretically justifiable to decode tokens in parallel using independent marginals, despite the true joint distribution potentially containing dependencies?} We address this with the following formal result, which characterizes the conditions under which greedy parallel (product of marginal distribution) decoding is equivalent to greedy sequential (true joint distribution) decoding in the high-confidence regime, and quantifies the divergence between the two distributions.

Prior to presenting the theorem, we will define the mathematical notation used in its statement. Let $p_{\boldsymbol{\theta}}(\cdot | E)$ denote the conditional probability mass function (PMF) given by an MDM condition on $E$ (comprising a prompt $p_0$ and previously generated tokens). Suppose the model is to predict $n$ tokens for positions $i_1, \dots, i_n$ not in $E$.
Let $\bfX = (X_{i_1}, \dots, X_{i_n})$ be the vector of $n$ tokens, where each $X_{i_j}$ takes values in vocabulary $\calV$.
Let $p(\bfX | E) \equiv p_{\boldsymbol{\theta}}(X_{i_1}, \dots, X_{i_n} | E)$ be the joint conditional PMF according to the model.
Let $p_j(X_{i_j} | E) \equiv p_{\boldsymbol{\theta}}(X_{i_j} | E)$ be the marginal conditional PMF for position $i_j$.
Parallel decoding generates tokens using the product of marginals: $q(\bfX | E) = \prod_{j=1}^n p_j(X_{i_j} | E)$. The proof of Theorem~\ref{thm:parallel decoding} and relevant discussions are in Appendix~\ref{sec:app proof}.
\begin{theorem}[Parallel Decoding under High Confidence]\label{thm:parallel decoding}
Suppose there exists a specific sequence of tokens $\bfx^* = (x_{i_1}, \dots, x_{i_n})$ such that for each $j \in \{1, \dots, n\}$, the model has high confidence in $x_{i_j}$: $p_j(X_{i_j} = x_{i_j} | E) > 1-\epsilon$ for some small $\epsilon > 0$. Then, the following results hold:

1. Equivalence for Greedy Decoding:
    If $(n+1)\epsilon \leq 1$ (i.e., $\epsilon \leq \frac{1}{n+1}$), then
    \begin{equation}
        \argmax_{\bfz} p(\bfz | E) = \argmax_{\bfz} q(\bfz | E) = \bfx^*.
    \end{equation}
    
This means that greedy parallel decoding (selecting $\argmax q$) yields the same result as greedy sequential decoding (selecting $\argmax p$).

This bound is tight: if $\epsilon > \frac{1}{n+1}$, there exist distributions $p(\bfX|E)$ satisfying the high-confidence marginal assumption for which $\argmax_{\bfz} p(\bfz | E) \neq \argmax_{\bfz} q(\bfz | E)$.

2. Distance and Divergence Bounds:
    Let $p(\cdot|E)$ and $q(\cdot|E)$ be denoted as $p$ and $q$ for brevity.

$L_p$ Distance ($p \ge 1$):
For $n>1$, $\LpDist{p}{p}{q} < ((n-1)^p+2n)^{1/p}\epsilon$.
Specifically, for Total Variation Distance ($D_{TV}(p,q) = \frac{1}{2}\LpDist{1}{p}{q}$): $D_{TV}(p,q) < \frac{3n-1}{2}\epsilon$.

Forward KL Divergence:
For $n>1$, $\KLDiv{p}{q} < (n-1)(H_b(\epsilon) + \epsilon \ln(|\calV|-1))$,
where $H_b(\epsilon) = -\epsilon\ln\epsilon-(1-\epsilon)\ln(1-\epsilon)$ is the binary entropy function, and $|\calV|$ is the size of the vocabulary.
\end{theorem}

\vspace{0.5em}
\noindent
Building on this theorem, we propose a practical \textit{factor}-based parallel decoding strategy as an extension of the threshold strategy that adaptively selects how many tokens to decode in parallel based on the confidence levels. Concretely, given the model's marginal confidence estimates for $n$ tokens in a block, we sort these confidences and select the largest $n$ such that $(n+1)(1 - c^{(n)}) < f$, where $f$ is a fixed decoding factor hyperparameter and $c^{(n)}$ is the $n$-th highest confidence. At each step, the top-$n$ tokens are decoded in parallel. This formulation mirrors the bound in Theorem~\ref{thm:parallel decoding} and ensures that decoding only proceeds when the marginal confidence is sufficiently high to approximate the joint decoding reliably. In contrast to the static threshold-based strategy, factor-based decoding dynamically controls the degree of parallelism in a theoretically grounded manner.

\begin{algorithm*}
\caption{Block-wise Confidence-aware Parallel Decoding with (Dual) KV Cache}
\begin{algorithmic}[1]
\Require $p_{\boldsymbol{\theta}}$, prompt $p_0$, answer length $L$, blocks $K$, block size $B$, steps per block $T$, threshold $\tau$, \texttt{use\_DualCache}, strategy $\in \{$\texttt{threshold}, \texttt{factor}$\}$, factor $f$
\State $x \gets [p_0; \texttt{[MASK]},..., \texttt{[MASK]}]$
\State \textbf{Initialize KV Cache} (single or dual) for $x$ (fuse with decoding). \hfill \textcolor{blue}{\textit{// KV Cache Init}}
\For{$k = 1$ to $K$}
    \State $s \gets |p_0| + (k-1)B$, \ $e \gets |p_0| + kB$
    \For{$t = 1$ to $T$}
        \State Use cache, run $p_{\boldsymbol{\theta}}$ on $x^{[s,e)}$ if \texttt{use\_DualCache} else $x^{[s,:)}$ \hfill \textcolor{blue}{\textit{// Cache Reuse}}
        
        \State For masked $x^i$, compute confidence $c^i = \max_x p_{\boldsymbol{\theta}}(x^i|\cdot)$ \hfill \textcolor{orange}{\textit{// Confidence scoring}}

        \If{\texttt{strategy} == \texttt{threshold}}
            \State Unmask all $i$ in $[s,e)$ with $c^i \geq \tau$, always unmask max $c^i$
        \ElsIf{\texttt{strategy} == \texttt{factor}}
            \State Sort $c^i$ in descending order as $(c^{(1)}, c^{(2)}, ..., c^{(m)})$
            \State Find largest $n$ such that $(n+1)(1 - c^{(n)}) < f$
            \State Unmask top-$n$ tokens, always unmask the max $c^i$
        \EndIf

        \If{all $x^{[s,e)}$ unmasked}
            \State \textbf{break}
        \EndIf
    \EndFor
    \State \textbf{Update KV cache}: 
    \textbf{if} \texttt{use\_DualCache}\textbf{:} prefix~\&~suffix; \textbf{else:} prefix. \hfill \textcolor{blue}{\textit{// Cache Update}}
\EndFor
\State \Return $x$
\end{algorithmic}
\label{alg:pipeline}
\end{algorithm*}
\section{Experiments}

\begin{table*}[t]
\centering
\caption{
Comprehensive benchmark results on the LLaDA-Instruct suite. Each cell presents the accuracy and the decoding throughput in tokens per second with relative speedup to the LLaDA baseline (bottom row, \textcolor{blue}{blue: tokens per second}/\textcolor{orange}{orange: relative speedup}). The highest throughput and speedup for each configuration are highlighted.}
\renewcommand{\arraystretch}{1.16}
\setlength{\tabcolsep}{3pt}
\footnotesize
\begin{tabular}{lc!{\vrule width 1pt}cccc}
\toprule
Benchmark & Gen Length & LLaDA & +Cache & +Parallel & +Cache+Parallel (\method) \\
\midrule
\multirow{2}{*}[-1.8ex]{\centering GSM8K (5-shot)}
    & 256 & 79.3 & 79.5 & 79.2 & 78.5 \\
    &       & \textcolor{blue}{6.7} (\textcolor{orange}{1$\times$}) 
            & \textcolor{blue}{21.2} (\textcolor{orange}{3.2$\times$}) 
            & \textcolor{blue}{16.5} (\textcolor{orange}{2.5$\times$}) 
            & \cellcolor{yellow!20}\textcolor{blue}{54.4} (\textcolor{orange}{8.1$\times$}) \\
    & 512 & 77.5 & 77.0 & 77.6 & 77.2 \\
    &       & \textcolor{blue}{3.2} (\textcolor{orange}{1$\times$}) 
            & \textcolor{blue}{10.4} (\textcolor{orange}{3.3$\times$}) 
            & \textcolor{blue}{18.6} (\textcolor{orange}{5.8$\times$}) 
            & \cellcolor{yellow!20}\textcolor{blue}{35.3} (\textcolor{orange}{11.0$\times$}) \\
\midrule
\multirow{2}{*}[-1.8ex]{\centering MATH (4-shot)}
    & 256 & 33.5 & 33.3 & 33.4 & 33.2 \\
    &       & \textcolor{blue}{9.1} (\textcolor{orange}{1$\times$})   
            & \textcolor{blue}{23.7} (\textcolor{orange}{2.6$\times$})
            & \textcolor{blue}{24.8} (\textcolor{orange}{2.7$\times$})
            & \cellcolor{yellow!20}\textcolor{blue}{51.7} (\textcolor{orange}{5.7$\times$})\\
    & 512 & 37.2 & 36.2 & 36.8 & 36.0 \\
    &       & \textcolor{blue}{8.0} (\textcolor{orange}{1$\times$})
            & \textcolor{blue}{19.7} (\textcolor{orange}{2.5$\times$})
            & \textcolor{blue}{23.8} (\textcolor{orange}{3.0$\times$})
            & \cellcolor{yellow!20}\textcolor{blue}{47.1} (\textcolor{orange}{5.9$\times$})\\
\midrule
\multirow{2}{*}[-1.8ex]{\centering HumanEval (0-shot)}
    & 256 & 41.5 & 42.7 & 43.9 & 43.3 \\
    &       & \textcolor{blue}{30.5} (\textcolor{orange}{1$\times$})   
            & \textcolor{blue}{40.7} (\textcolor{orange}{1.3$\times$})
            & \textcolor{blue}{101.5} (\textcolor{orange}{3.3$\times$})
            & \cellcolor{yellow!20}\textcolor{blue}{114.1} (\textcolor{orange}{3.7$\times$})\\
    & 512 & 43.9 & 45.7 & 43.3 & 44.5 \\
    &       & \textcolor{blue}{18.4} (\textcolor{orange}{1$\times$})
            & \textcolor{blue}{29.3} (\textcolor{orange}{1.6$\times$})
            & \textcolor{blue}{57.1} (\textcolor{orange}{3.1$\times$})
            & \cellcolor{yellow!20}\textcolor{blue}{73.7} (\textcolor{orange}{4.0$\times$})\\
\midrule
\multirow{2}{*}[-1.8ex]{\centering MBPP (3-shot)}
    & 256 & 29.4 & 29.6 & 28.4 & 28.2 \\ 
    &       & \textcolor{blue}{6.0} (\textcolor{orange}{1$\times$})   
            & \textcolor{blue}{17.0} (\textcolor{orange}{2.8$\times$})
            & \textcolor{blue}{24.8} (\textcolor{orange}{4.1$\times$})
            & \cellcolor{yellow!20}\textcolor{blue}{44.8} (\textcolor{orange}{7.5$\times$})\\
    & 512 & 14.8 & 13.4 & 15.0 & 13.8   \\
    &       & \textcolor{blue}{4.3} (\textcolor{orange}{1$\times$})
            & \textcolor{blue}{10.1} (\textcolor{orange}{2.3$\times$})
            & \textcolor{blue}{22.3} (\textcolor{orange}{5.1$\times$})
            & \cellcolor{yellow!20}\textcolor{blue}{39.5} (\textcolor{orange}{9.2$\times$})\\
\bottomrule
\label{tab:llada_main}
\end{tabular}
\vspace{-10pt}
\end{table*}
\begin{table*}[t]
\centering
\caption{
Comprehensive benchmark results on Dream-Base variants over four tasks with different generation lengths (256 and 512). 
Each cell shows accuracy (top row) and decoding throughput in tokens per second with relative speedup to Dream-Base baseline (bottom row, \textcolor{blue}{blue: tokens per second}/\textcolor{orange}{orange: relative speedup}). 
Numbers in \cellcolor{yellow!20}yellow indicate the highest throughput and speedup per configuration. 
}
\renewcommand{\arraystretch}{1.15}
\setlength{\tabcolsep}{4pt}
\footnotesize
\begin{tabular}{lc!{\vrule width 1pt}cccc}
\toprule
Benchmark & Gen Length & Dream & +Cache & +Parallel & +Cache+Parallel (\method) \\
\midrule
\multirow{2}{*}[-1.8ex]{\centering GSM8K (5-shot)}
    & 256 & 75.0 & 74.3 & 74.2 & 74.8 \\
    &         & \textcolor{blue}{9.1} (\textcolor{orange}{1$\times$}) 
              & \textcolor{blue}{32.5} (\textcolor{orange}{3.6$\times$}) 
              & \textcolor{blue}{14.2} (\textcolor{orange}{1.6$\times$})
              & \cellcolor{yellow!20}\textcolor{blue}{48.2} (\textcolor{orange}{5.3$\times$}) \\
\multirow{2}{*}{} 
    & 512 & 76.0 & 74.3 & 73.4 & 74.0 \\
    &         & \textcolor{blue}{7.7} (\textcolor{orange}{1$\times$}) 
              & \textcolor{blue}{25.6} (\textcolor{orange}{3.3$\times$}) 
              & \textcolor{blue}{14.6} (\textcolor{orange}{1.9$\times$})
              & \cellcolor{yellow!20}\textcolor{blue}{42.9} (\textcolor{orange}{5.6$\times$}) \\
\midrule
\multirow{2}{*}[-1.8ex]{\centering MATH (4-shot)}
    & 256 & 38.4 & 36.8 & 37.9 & 37.6 \\
    &         & \textcolor{blue}{11.4} (\textcolor{orange}{1$\times$}) 
              & \textcolor{blue}{34.3} (\textcolor{orange}{3.0$\times$}) 
              & \textcolor{blue}{27.3} (\textcolor{orange}{2.4$\times$})
              & \cellcolor{yellow!20}\textcolor{blue}{66.8} (\textcolor{orange}{5.9$\times$}) \\
\multirow{2}{*}{}
    & 512 &    39.8  & 38.0 & 39.5 & 39.3 \\
    &         & \textcolor{blue}{9.6} (\textcolor{orange}{1$\times$})
              & \textcolor{blue}{26.8} (\textcolor{orange}{2.8$\times$})
              & \textcolor{blue}{31.6} (\textcolor{orange}{3.2$\times$})
              & \cellcolor{yellow!20}\textcolor{blue}{63.3} (\textcolor{orange}{6.5$\times$}) \\
\midrule
\multirow{2}{*}[-1.8ex]{\centering HumanEval (0-shot)}
    & 256 & 49.4 & 53.7 & 49.4 & 54.3 \\
    &         & \textcolor{blue}{23.3} (\textcolor{orange}{1$\times$}) 
              & \textcolor{blue}{35.2} (\textcolor{orange}{1.5$\times$}) 
              & \textcolor{blue}{45.6} (\textcolor{orange}{2.0$\times$})
              & \cellcolor{yellow!20}\textcolor{blue}{62.0} (\textcolor{orange}{2.8$\times$}) \\
\multirow{2}{*}{}
    & 512 & 54.3 & 54.9 & 51.8 & 54.3 \\
    &         & \textcolor{blue}{16.3} (\textcolor{orange}{1$\times$})
              & \textcolor{blue}{27.8} (\textcolor{orange}{1.7$\times$})
              & \textcolor{blue}{29.8} (\textcolor{orange}{1.8$\times$})
              & \cellcolor{yellow!20}\textcolor{blue}{52.8} (\textcolor{orange}{3.2$\times$}) \\
\midrule
\multirow{2}{*}[-1.8ex]{\centering MBPP (3-shot)}
    & 256 & 56.6 & 53.2 & 53.8 & 56.4 \\
    &         & \textcolor{blue}{11.2} (\textcolor{orange}{1$\times$}) 
              & \textcolor{blue}{34.5} (\textcolor{orange}{3.1$\times$}) 
              & \textcolor{blue}{31.8} (\textcolor{orange}{2.8$\times$})
              & \cellcolor{yellow!20}\textcolor{blue}{76.0} (\textcolor{orange}{6.8$\times$}) \\
\multirow{2}{*}{}
    & 512 & 55.6 & 53.8 & 55.4 & 55.2 \\
    &         & \textcolor{blue}{9.4} (\textcolor{orange}{1$\times$})
              & \textcolor{blue}{26.7} (\textcolor{orange}{2.8$\times$})
              & \textcolor{blue}{37.6} (\textcolor{orange}{4.0$\times$})
              & \cellcolor{yellow!20}\textcolor{blue}{73.6} (\textcolor{orange}{7.8$\times$}) \\
\bottomrule
\end{tabular}
\label{tab:dream_main}
\vspace{-10pt}
\end{table*}

\subsection{Experimental Setup}
All experiments are conducted on an NVIDIA A100 80GB GPU. The proposed approach, \method, comprises two components: a Key-Value Cache mechanism and a Confidence-Aware Parallel Decoding strategy. The KV Cache component introduces a hyperparameter, the cache block size, varied between 4 and 32. The parallel decoding strategy uses a confidence threshold hyperparameter, explored in the range of 0.5 to 1.0. Unless otherwise specified, we use PrefixCache with block size of 32 and the threshold to 0.9.

\begin{wrapfigure}{r}{0.42\textwidth}
    \centering
    \includegraphics[width=\linewidth]{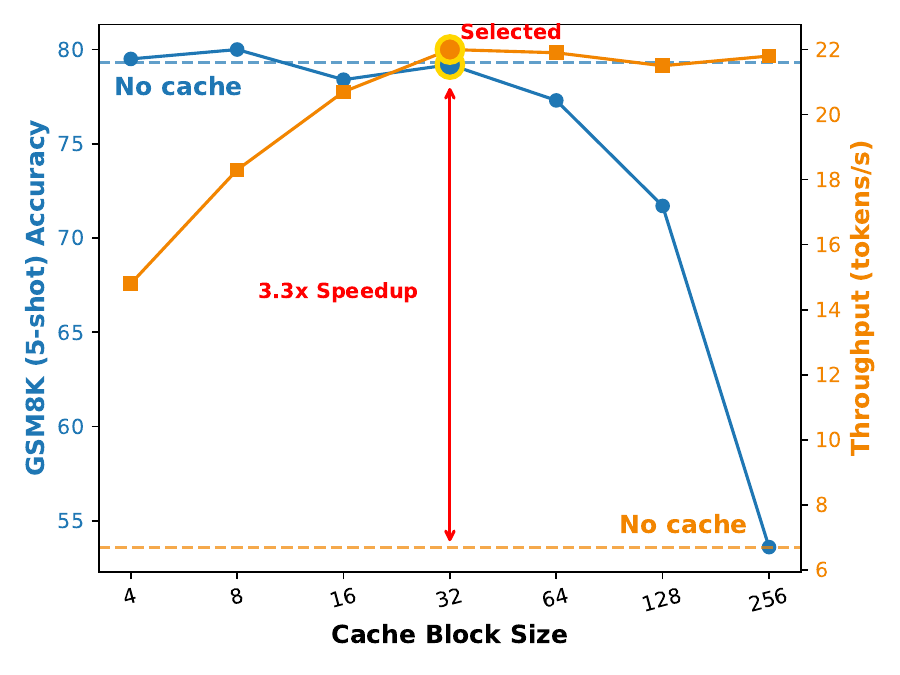}
    \caption{\textbf{Impact of Cache Block Size on Accuracy and Throughput.} The orange line illustrates the effect of varying cache block size on throughput, while the blue line depicts accuracy.}
    \label{fig:cache-block-size}
    \vspace{-20pt}
\end{wrapfigure}

We evaluate \method on two recent diffusion-based language models: LLaDA~\cite{nie2025largelanguagediffusionmodels}, LLaDA-1.5~\cite{zhu2025llada15variancereducedpreference} and Dream~\cite{dream2025}. Benchmarks include four widely-used datasets—GSM8K, MATH, HumanEval, and MBPP, to assess performance across diverse reasoning and code generation tasks. We also test under varying generation lengths to evaluate scalability and robustness.

In addition, we extend our evaluation to LLaDA-V~\cite{you2025llada}, a multimodal variant of LLaDA tailored for vision-language reasoning tasks. For this, we use two challenging multimodal benchmarks: MathVista and MathVerse, which require solving math problems grounded in complex visual scenes.

Inference throughput is measured as the average number of output tokens generated per second, calculated over the full sequence until the end-of-sequence (\texttt{<eos>}) token is reached. This metric reflects true end-to-end decoding speed. All evaluations are conducted using the standardized \texttt{lm-eval} library to ensure consistency and reproducibility.

\subsection{Main Results: Performance and Speed}

We report decoding performance and efficiency gains for \method on both the LLaDA-Instruct and Dream-Base models across the four benchmarks in Tables~\ref{tab:llada_main} and~\ref{tab:dream_main}.

Overall, introducing the KV Cache mechanism yields significant speed improvements for all tasks and sequence lengths, typically achieving a $2\times$ to $3.6\times$ speedup compared to the vanilla backbone. When the parallel decoding strategy is applied individually, we see additional acceleration, often pushing speedups to $4\times$–$6\times$ for the evaluated settings, particularly as the generation length increases.

When both techniques are combined, the improvements become even more pronounced. On LLaDA, for example, combined KV Cache and parallel decoding methods boost throughput by up to $11\times$ (GSM8K, length 512) and $9.2\times$ (MBPP, length 512) over the standard baseline. Similarly, on Dream-Base, the largest throughput gains are observed on MBPP ($7.8\times$ at length 512) and GSM8K ($5.6\times$ at length 512). These results indicate that not only are our methods effective individually, but they are also highly complementary, resulting in the combined acceleration.

\begin{figure*}[h]
    \centering
    \includegraphics[width=\textwidth]{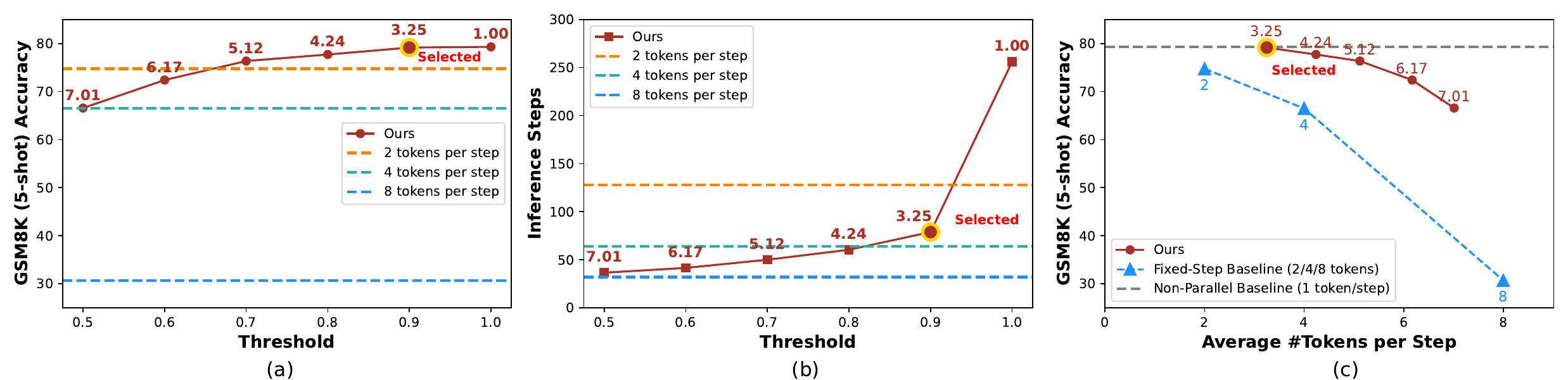} 
    \caption{
    (a) The red line shows the GSM8K (5-shot) accuracy across different confidence thresholds. Numbers along the red line indicate the average number of tokens decoded at each step. The three dashed lines represent the accuracy of the baseline method when selecting the top 2, 4, or 8 tokens per step.
(b) The number of inference steps required under varying confidence thresholds.
(c) A comparison between our method and the baseline on GSM8K (5-shot) accuracy, plotted against the average number of tokens per step. Our method consistently outperforms the baseline.
    }
    \label{fig:threshold-metrics}
    \vspace{-10pt}
\end{figure*}

Importantly, these efficiency gains are achieved with negligible impact on accuracy. Across all benchmarks and settings, the accuracy of our accelerated methods remains within 1–2 points of the backbone, and in several cases, accuracy is even slightly improved. This demonstrates that the speedup comes at almost no cost to task performance, ensuring reliability for practical deployment. We also observe that longer sequences, which are common in few-shot and code generation scenarios, benefit proportionally more from our caching and parallelization techniques due to greater opportunities for cache reuse and batch computation. We also evaluate an advanced version, LLaDA-1.5, which achieves consistently stronger accuracy and comparable or higher throughput across benchmarks (Table~\ref{tab:llada_1.5}).

In addition to text-only models, we evaluate \method on the multimodal LLaDA-V using the MathVista and MathVerse datasets, which require complex vision-language reasoning. As shown in Table~\ref{tab:blocklen_effect}, LLaDA-V shows a strong sensitivity to block size, with accuracy dropping by over 8\% when reducing from 96 to 8 on MathVista. To address this, we retain a full block length and apply refresh-based updates instead of small-block caching. This yields up to $9.9\times$ speedup with minimal accuracy degradation (Table~\ref{tab:lladav_main_results}). On MathVerse, accuracy is even slightly improved under Fast-dLLM, demonstrating the broad applicability of our method to multimodal reasoning tasks.

Furthermore, the improvements generalize across model architectures (LLaDA and Dream), task types (math reasoning, program synthesis), and modalities (text and vision), confirming that \method is a practical and broadly applicable framework for accelerating masked diffusion-based language models.

\begin{table}[t]
\centering
\small
\caption{\textbf{Performance and Speedup Comparison of LLaDA-V on MathVista and MathVerse.}
Each benchmark includes results from Full Steps, Half Steps, and \method. \method significantly improves throughput (highlighted), with minimal accuracy loss.}
\setlength{\tabcolsep}{6pt}
\begin{tabular}{l|ccc|ccc}
\toprule
\multirow{2}{*}{\textbf{Metric}} & \multicolumn{3}{c|}{\textbf{MathVista}} & \multicolumn{3}{c}{\textbf{MathVerse}} \\
\cmidrule(lr){2-4} \cmidrule(lr){5-7}
& Full Steps & Half Steps & \method & Full Steps & Half Steps & \method \\
\midrule
Accuracy (\%) 
& 59.2 & 59.7 & 56.6 
& 28.5 & 28.3 & 28.6 \\
Throughput (Speedup)
& \textcolor{blue}{2.84} (\textcolor{orange}{1×}) 
& \textcolor{blue}{5.56} (\textcolor{orange}{1.96×}) 
& \cellcolor{yellow!20}\textcolor{blue}{28.2} (\textcolor{orange}{9.9×}) 
& \textcolor{blue}{2.75} (\textcolor{orange}{1×}) 
& \textcolor{blue}{5.17} (\textcolor{orange}{1.88×}) 
& \cellcolor{yellow!20}\textcolor{blue}{23.3} (\textcolor{orange}{8.5×}) \\
\bottomrule
\end{tabular}
\label{tab:lladav_main_results}
\end{table}

\subsection{Ablations and Analysis}
\begin{table*}[htb]
\centering
\begin{minipage}[t]{0.48\textwidth}
    \centering
    \scriptsize
    \captionof{table}{\textbf{Performance and Speedup Comparison on LLaDA Between 5-Shot and 8-Shot Settings at Generation Length 1024.}
This table compares the accuracy and throughput speedups of different decoding strategies under 5-shot and 8-shot configurations using a generation length of 1024. The results demonstrate how increased prefill length enhances the effectiveness of caching strategies, particularly for DualCache.}
    \setlength{\tabcolsep}{5pt}
    \begin{tabular}{l|c|ccc}
    \toprule
        \multirow{2}{*}{Setting.} & \multirow{2}{*}{LLaDA} & \multicolumn{3}{c}{Parallel Decoding} \\
                          &                        & No Cache & PrefixCache & DualCache \\
    \midrule
    \multirow{2}{*}{5-shot}
      & 77.0 & 77.4 & 75.2 & 74.7\\
      & \textcolor{blue}{1.1} (\textcolor{orange}{1×}) 
      & \textcolor{blue}{11.7} (\textcolor{orange}{10.6×}) 
      & \textcolor{blue}{14.4} (\textcolor{orange}{13.1×}) 
      & \cellcolor{yellow!20}\textcolor{blue}{21.6} (\textcolor{orange}{19.6×}) \\
    \multirow{2}{*}{8-shot}
      & 77.3 & 78.0 & 75.7 & 76.0 \\
      & \textcolor{blue}{0.7} (\textcolor{orange}{1×}) 
      & \textcolor{blue}{9.3} (\textcolor{orange}{13.3×}) 
      & \textcolor{blue}{13.0} (\textcolor{orange}{18.6×}) 
      & \cellcolor{yellow!20}\textcolor{blue}{19.3} (\textcolor{orange}{27.6×}) \\
    \bottomrule
    \end{tabular}
    \label{tab:5v8shot}
\end{minipage}
\hfill
\begin{minipage}[t]{0.48\textwidth}
    \centering
    \scriptsize
    \captionof{table}{\textbf{Impact of Generation Length on Accuracy and Speedup Under 8-Shot for LLaDA.}
This table illustrates the effect of varying generation lengths (256, 512, and 1024) on decoding performance and efficiency for different caching strategies under the 8-shot setting. Longer generation lengths lead to higher throughput gains, especially for DualCache, validating the scalability of our approach.}
    \setlength{\tabcolsep}{5pt}
    \begin{tabular}{l|c|ccc}
    \toprule
    \multirow{2}{*}{Len.} & \multirow{2}{*}{LLaDA} & \multicolumn{3}{c}{Parallel Decoding} \\
                          &                        & No Cache & PrefixCache & DualCache \\
    \midrule
    \multirow{2}{*}{256}
        & 77.6 & 77.9 & 77.3 & 76.9 \\
        & \textcolor{blue}{4.9} (\textcolor{orange}{1×}) 
          & \textcolor{blue}{16.4} (\textcolor{orange}{3.3×}) 
          & \cellcolor{yellow!20}\textcolor{blue}{49.2} (\textcolor{orange}{10.0×}) 
          & \textcolor{blue}{46.3} (\textcolor{orange}{9.4×}) \\
    \multirow{2}{*}{512}
        & 78.9 & 78.9 & 74.8 & 75.4 \\
        & \textcolor{blue}{2.3} (\textcolor{orange}{1×}) 
          & \textcolor{blue}{14.0} (\textcolor{orange}{6.1×}) 
          & \textcolor{blue}{32.0} (\textcolor{orange}{13.9×}) 
          & \cellcolor{yellow!20}\textcolor{blue}{36.4} (\textcolor{orange}{15.8×}) \\
    \multirow{2}{*}{1024}
        & 77.3 & 78.0 & 75.7 & 76.0 \\
        & \textcolor{blue}{0.7} (\textcolor{orange}{1×}) 
          & \textcolor{blue}{9.3} (\textcolor{orange}{13.3×}) 
          & \textcolor{blue}{13.0} (\textcolor{orange}{18.6×}) 
          & \cellcolor{yellow!20}\textcolor{blue}{19.3} (\textcolor{orange}{27.6×}) \\
    \bottomrule
    \end{tabular}
    \label{tab:genlength}
\end{minipage}
\end{table*}

We conduct extensive ablation studies to understand how different components of \method contribute to performance, focusing on factors such as prefill length, generation length, cache mechanism variants, cache block size, and confidence thresholds.

\paragraph{Influence of Prefill and Generation Length on Acceleration} Table~\ref{tab:5v8shot} and Table~\ref{tab:genlength} indicate that both prefill length (\emph{n}-shot) and generation length markedly impact overall speedup. Specifically, as the prefill length increases from 5-shot to 8-shot, the speedup obtained by both versions of KV Cache rises significantly (e.g., speedup for DualCache increases from 19.6$\times$ in 5-shot to 27.6$\times$ in 8-shot for generation length 1024). Similarly, extending the generation length amplifies the potential for cache reuse, leading to higher speedup. Notably, for 8-shot, speedup with DualCache grows from 9.4$\times$ (gen len 256) up to 27.6$\times$ (gen len 1024). This aligns with the theoretical expectation that amortizing computation over longer sequences yields more pronounced efficiency gains.

\paragraph{Comparison of prefix KV Cache vs.\ DualCache}
We further compare our prefix KV Cache and DualCache versions in multiple settings. As shown in Table~\ref{tab:genlength}, DualCache generally achieves higher speedup than the prefix KV Cache, especially for longer generation lengths. For gen len 512 and 1024, DualCache demonstrates up to 27.6$\times$ speedup, outperforming the prefix KV Cache's 18.6$\times$ in the same scenario. Importantly, DualCache maintains competitive accuracy, with only minor trade-offs relative to the cache-only variant. This highlights DualCache's effectiveness in exploiting parallelism and cache locality for both efficiency and accuracy.

\paragraph{Effect of Cache Block Size}
Figure~\ref{fig:cache-block-size} analyzes the influence of the cache block size hyperparameter. We observe that smaller block sizes tend to maximize accuracy but incur overhead due to frequent cache updates. In contrast, larger block sizes may diminish accuracy owing to increased context mismatch. Block size of 32 achieves the best trade-off, substantially improving throughput while largely preserving accuracy. This hyperparameter thus offers a practical knob for balancing latency and precision in real deployments.

\paragraph{Dynamic Threshold vs.\ Fixed Token-per-Step Strategies}
We evaluate our Confidence-Aware Parallel Decoding method against fixed token-per-step baselines on GSM8K (Figure~\ref{fig:threshold-metrics}). Our adaptive strategy consistently outperforms fixed baselines across key metrics: it delivers higher accuracy at comparable or reduced number of function evaluations (NFE) and generates more tokens per step on average while closely tracking accuracy. In the rightmost panel, the dynamic method approaches or exceeds the accuracy of the 1-token (non-parallel) baseline, but with much greater throughput. The result demonstrates the effectiveness of Confidence-Aware Parallel Decoding, offering practical advantages.

\paragraph{Factor Decoding vs.\ Fixed Token-per-Step Strategies}
We further compare our factor-based parallel decoding approach with fixed token-per-step baselines on GSM8K (Figure~\ref{fig:factor-metrics}) and with the threshold-based strategy (Table~\ref{tab:threshold-vs-factor}). Across a range of factor values, our method consistently achieves competitive or higher accuracy with fewer inference steps. As the factor increases, the number of tokens decoded per step grows steadily, reducing iteration count while maintaining performance. Compared to the threshold strategy, factor decoding achieves similar accuracy but significantly higher throughput by adaptively controlling decoding granularity. We also analyze parallel token counts across decoding step at Appendix~\ref{subsec:token-count-stats}.

\paragraph{Decoding Efficiency Analysis and Limitations}
As discussed in Section~\ref{subsection:throughput-comparison}, PrefixCache significantly accelerates diffusion-based LLMs like LLaDA with up to $5\times$ throughput improvement in compute-bound scenarios compared to LLaDA. At smaller batch sizes, PrefixCache achieves throughput comparable to or even exceeding that of autoregressive models like LLaMA. However, as batch sizes grow, PrefixCache struggles to match LLaMA, which transitions from memory-bound to compute-bound performance. This reflects a general challenge for diffusion-based LLMs, which tend to incur higher computational overhead due to full attention operations during decoding.

\section{Related Work}
\subsection{Diffusion LLM}
Diffusion models have emerged as a transformative paradigm in generative modeling, initially achieving remarkable success in continuous domains such as image~\cite{rombach2022highresolutionimagesynthesislatent, nichol2022glidephotorealisticimagegeneration, ramesh2021zeroshottexttoimagegeneration,saharia2022photorealistictexttoimagediffusionmodels} and audio synthesis~\cite{yang2023diffsounddiscretediffusionmodel, huang2023makeanaudiotexttoaudiogenerationpromptenhanced} before expanding into natural language processing. Recent advancements in discrete diffusion models~\cite{austin2021structured,nie2025scalingmaskeddiffusionmodels, nie2025largelanguagediffusionmodels,hoogeboom2021argmax,campbell2022continuous,he2022diffusionbert,meng2022concrete,reid2022diffuser,sun2022score,kitouni2023disk,zheng2023judging,chen2023fast,ye2023diffusion,sahoo2024simple,shi2024simplified,zheng2024masked,gat2024discrete,yu2025dimplediscretediffusionmultimodal,yu2025discretediffusionlargelanguage} have reshaped the landscape of text generation, offering a viable alternative to autoregressive (AR) paradigms in large language models (LLMs). These models address the inherent challenges of discrete data by redefining noise injection and denoising processes through innovative mathematical formulations.

\noindent\textbf{Theoretical Foundations of Discrete Diffusion} Diffusion models for discrete data were first explored in~\cite{sohl2015deep,hoogeboom2021argmax}.
Subsequently, D3PM~\cite{austin2021structured} provided a more general framework. This framework models the forward noising process as a discrete state Markov chain using specific transition matrices. For the reverse process, D3PM learns a parameterized model of the conditional probability of the original data given a noised version by maximizing the Evidence Lower Bound (ELBO).
CTMC~\cite{campbell2022continuous} further extended D3PM to a continuous-time setting, formalizing it as a continuous-time Markov Chain (CTMC).
In a distinct approach, SEDD~\cite{lou2023discrete} learns the reverse process by parameterizing the ratio of marginal likelihoods for different data instances at a given noising timestep. This ratio model is then trained using a Denoising Score Entropy objective. More recently, research on Masked Diffusion Models (MDMs) by MDLM~\cite{shi2024simplified,sahoo2024simple,zheng2024masked} and RADD~\cite{ou2024your} has introduced significant clarifications. These studies have demonstrated that different parameterizations of MDMs can be equivalent.

\noindent\textbf{Integration with Pre-trained Language Models} A critical breakthrough involves combining discrete diffusion with existing LLM architectures. Diffusion-NAT~\cite{zhou2023diffusionnatselfpromptingdiscretediffusion} unifies the denoising process of discrete diffusion with BART’s~\cite{lewis2019bartdenoisingsequencetosequencepretraining} non-autoregressive decoding, enabling iterative refinement of masked tokens. By aligning BART’s inference with diffusion steps, this approach leverages pre-trained knowledge while maintaining generation speed 20× faster than comparable AR transformers. Similarly, the LLaDA~\cite{nie2025largelanguagediffusionmodels} and DiffuLLaMA~\cite{gong2024scaling} framework scales diffusion to $7$B parameters using masked denoising, while LLaDA and Dream~\cite{dream2025} demonstrating competitive performance with autoregressive baselines like LLaMA3~\cite{grattafiori2024llama3herdmodels} through recursive token prediction across diffusion timesteps.
\subsection{LLM Acceleration}
\noindent\textbf{Key-Value Cache.}
Key-Value (KV) Cache is a fundamental optimization technique in modern large language model (LLM) inference with Transformer architecture~\cite{vaswani2017attention}. It enables efficient autoregressive text generation by storing and reusing previously computed attention states. However, it is non-trival to apply KV Cache in diffusion langauge models such as LLaDA due to full attention. Block diffusion~\cite{arriola2025blockdiffusioninterpolatingautoregressive} overcomes key limitation of previous diffusion langauge models by generating block-by-block so that key and values of previously decoded blocks can be stored and reused. 

\noindent\textbf{Non-Autoregressive Generation}
Non-autoregressive (NAR) generation marks a fundamental shift from sequential token generation by enabling the simultaneous generation of multiple tokens, significantly accelerating inference~\cite{xiao2023surveynonautoregressivegenerationneural}. Initially introduced for neural machine translation, NAR methods have since been extended to a variety of tasks, including grammatical error correction, text summarization, dialogue systems, and automatic speech recognition. Although NAR generation offers substantial speed advantages over autoregressive approaches, it often sacrifices generation quality. Diffusion LLMs represent a recent paradigm for non-autoregressive text generation; however, prior work~\cite{nie2025largelanguagediffusionmodels} has struggled to realize the expected acceleration due to a notable drop in output quality.

\section{Conclusion}
In this work, we tackle key limitations in the inference efficiency of Diffusion-based Large Language Models (Diffusion LLMs), which have historically lacked support for KV Cache and exhibited performance degradation during parallel decoding. To bridge the gap with autoregressive models, we propose \method, a diffusion-based framework that introduces an approximate KV Cache mechanism tailored to the bidirectional attention characteristics of Diffusion LLMs, enabled by a block-wise generation scheme. Furthermore, we identify that the main obstacle to effective parallel decoding is the disruption of token dependencies arising from the conditional independence assumption. To address this, \method employs a Confidence-Aware Parallel Decoding strategy that facilitates safe and efficient multi-token generation. Extensive experiments across multiple benchmarks and model baselines (LLaDA and Dream) show that \method achieves up to a 27.6$\times$ speedup with minimal loss in accuracy. These findings offer a practical solution for deploying Diffusion LLMs as competitive alternatives to autoregressive models in real-world applications.

\newpage
\appendix
\onecolumn

\section{Proof}
\label{sec:app proof}
In this section, we will give the comprehensive proof and discussion of Theorem~\ref{thm:parallel decoding}.
\begin{proof}

\textbf{Step 1: Show that $\boldsymbol{x}^*$ is the unique maximizer of $q(x)$.}

Let $p_j^* = p_j(X_{i_j} = x_{i_j} | E)$. We are given $p_j^* > 1-\epsilon$.
Let $\epsilon'_j = 1 - p_j^* = p_j(X_{i_j} \neq x_{i_j} | E)$. Thus, $\epsilon'_j < \epsilon$. The product-of-marginals probability mass function (PMF) is
\begin{equation*}
q(\bfz | E) = \prod_{j=1}^n p_j(X_{i_j} = z_j | E).
\end{equation*}
To maximize $q(\bfz | E)$, we must maximize each term $p_j(X_{i_j} = z_j | E)$ independently. The condition $(n+1)\epsilon \leq 1$ implies $\epsilon \leq 1/(n+1)$. Since $n \ge 1$, it follows that $1/(n+1) \le 1/2$.
So, $\epsilon \leq 1/2$. Therefore, for the chosen $x_{i_j}$:
\begin{equation*}
p_j^* = p_j(X_{i_j} = x_{i_j} | E) > 1-\epsilon \geq 1-1/2 = 1/2.
\end{equation*}
This means $x_{i_j}$ is the unique maximizer for $p_j(\cdot | E)$.
So,
\begin{equation*}
\argmax_{\bfz} q(\bfz | E) = (x_{i_1}, \dots, x_{i_n}) = \bfx^*.
\end{equation*}

\textbf{Step 2: Show that $\boldsymbol{x}^*$ is the unique maximizer of $p(x)$.}

We want to show $p(\bfx^* | E) > p(\bfz | E)$ for all $\bfz \neq \bfx^*$.
Using the Bonferroni inequality:
\begin{align*}
p(\bfx^* | E) &= p(\cap_{j=1}^n \{X_{i_j} = x_{i_j}\} | E) \ge 1 - \sum_{j=1}^n p(X_{i_j} \neq x_{i_j} | E) = 1 - \sum_{j=1}^n \epsilon'_j.
\end{align*}
Since $\epsilon'_j < \epsilon$ for all $j$, we have $\sum_{j=1}^n \epsilon'_j < n\epsilon$.
So,
\begin{equation*}
p(\bfx^* | E) > 1 - n\epsilon.
\end{equation*}
Now consider any $\bfz = (z_1, \dots, z_n)$ such that $\bfz \neq \bfx^*$.
This means there is at least one index $k$ such that $z_k \neq x_{i_k}$.
The event $\{\bfX=\bfz\}$ is a sub-event of $\{X_{i_k}=z_k\}$.
So,
\begin{equation*}
p(\bfz | E) \le p_k(X_{i_k}=z_k | E).
\end{equation*}
Since $z_k \neq x_{i_k}$,
\begin{equation*}
p_k(X_{i_k}=z_k | E) \le p_k(X_{i_k} \neq x_{i_k} | E) = \epsilon'_k < \epsilon.
\end{equation*}
Thus,
\begin{equation*}
p(\bfz | E) < \epsilon.
\end{equation*}
For $p(\bfx^* | E) > p(\bfz | E)$ to hold, it is sufficient that
\begin{equation*}
1 - n\epsilon \geq \epsilon,
\end{equation*}
which simplifies to $1 \geq (n+1)\epsilon$, or $\epsilon \leq \frac{1}{n+1}$.
The theorem assumes $(n+1)\epsilon < 1$, which is exactly this condition.
The strict inequalities $p(\bfx^* | E) \geq 1 - \sum \epsilon'_j > 1 - n\epsilon$ and $p(\bfz | E) \le \epsilon'_k < \epsilon$ ensure that $p(\bfx^* | E) > p(\bfz | E)$.
Thus,
\begin{equation*}
\argmax_{\bfz} p(\bfz | E) = \bfx^*.
\end{equation*}
Combined with the argmax of $q$, this proves the main statement of Part 1:
\begin{equation*}
\argmax_{\bfz} p(\bfz | E) = \argmax_{\bfz} q(\bfz | E) = \bfx^*.
\end{equation*}

\textbf{Step 3: Tightness of the bound $\frac{1}{n+1}$.}

The bound $\epsilon \leq \frac{1}{n+1}$ is tight. This means if $\epsilon > \frac{1}{n+1}$, one can construct a scenario where the marginal conditions $p_j(X_{i_j}=x_{i_j}|E) > 1-\epsilon$ hold, but $\argmax_{\bfz} p(\bfz | E) \neq \bfx^*$ (which is $\argmax_{\bfz} q(\bfz | E)$ as long as $\epsilon \leq 1/2$).

Consider a vocabulary $\calV=\{0,1\}$ and let $x_{i_j}=0$ for all $j$, so $\bfx^* = (0, \dots, 0)$. For each $j \in \{1, \dots, n\}$, let $\mathbf{e}_j$ be the vector with $1$ at position $j$ and $0$ elsewhere. Let $\eta = \frac{1}{n+1}(\epsilon - \frac{1}{n+1}) > 0$. Set $p(\mathbf{e}_j|E) = \frac{1}{n+1} + \frac{1}{n}\eta, \ \forall 1 \leq j \leq n$ and $p(\bfx^*|E) = \frac{1}{n+1} - \eta$ , then $\bfx^* \notin \argmax_{\bfz} p(\bfz|E)$.
The marginal probabilities are:
\begin{align*}
p_j(X_{i_j}=1|E) &= p(\mathbf{e}_j|E)  = \frac{1}{n+1} + \frac{1}{n}\eta, \ \forall 1 \leq j \leq n.\\
p_j(X_{i_j}=0|E) &= 1 - p_j(X_{i_j}=1|E) = 1 - \epsilon_c = \frac{n}{n+1} - \frac{1}{n}\eta > 1-\epsilon,
\end{align*}
because
\begin{equation*}
\frac{1}{n}\eta = \frac{1}{n(n+1)}(\epsilon - \frac{1}{n+1}) < \epsilon - \frac{1}{n+1}
\end{equation*}
So, the marginal condition $p_j(X_{i_j}=x_{i_j}|E) > 1-\epsilon$ (with $x_{i_j}=0$) holds. As shown, $\argmax_{\bfz} p(\bfz|E)$ can be made different from $\bfx^*$. Thus, if $\epsilon > \frac{1}{n+1}$, the argmax of $p$ and $q$ may not be the same.

\textbf{Step 4: Bound the $L_p$ distance.}
Let $A_j$ be the event $\{X_{i_j}=x_{i_j}\}$.
\begin{equation*}
\LpDist{p}{p}{q}^p = |p(\bfx^*|E) - q(\bfx^*|E)|^p + \sum_{\bfz \neq \bfx^*} |p(\bfz|E) - q(\bfz|E)|^p.
\end{equation*}
The term $|p(\cap_{j=1}^n A_j|E) - \prod_{j=1}^n p(A_j|E)|$ (using $p(A_j|E)$ for $p_j(X_{i_j}=x_{i_j}|E)$) can be bounded. Since
\begin{equation*}
 1 - \sum_{j=1}^n \epsilon'_j\leq p(\cap_{j=1}^n A_j|E) \leq \min_{1\leq j\leq n} p(A_j|E) = 1 - \max_{1\leq j\leq n} \epsilon'_j,
\end{equation*}
\begin{equation*}
 1 - \sum_{j=1}^n \epsilon'_j  \leq \prod_{j=1}^n (1-\epsilon'_j) = \prod_{j=1}^n p(A_j|E) \leq 1 - \max_{1\leq j\leq n} \epsilon'_j.
\end{equation*}
Thus,
\begin{equation*}
|p(\bfx^*|E) - q(\bfx^*|E)| < (n-1)\epsilon.
\end{equation*}
For $\bfz \neq \bfx^*$: $p(\bfz|E) < \epsilon$ and $q(\bfz|E) < \epsilon$. So,
\begin{equation*}
|p(\bfz|E) - q(\bfz|E)| < \epsilon.
\end{equation*}
The sum $\sum_{\bfz \neq \bfx^*} |p(\bfz|E) - q(\bfz|E)|$ can be bounded:
\begin{align*}
\sum_{\bfz \neq \bfx^*} |p(\bfz|E) - q(\bfz|E)| &\le \sum_{\bfz \neq \bfx^*} (p(\bfz|E) + q(\bfz|E)) = p(\bfX \neq \bfx^*|E) + q(\bfX \neq \bfx^*|E).
\end{align*}
\begin{align*}
p(\bfX \neq \bfx^*|E) &= 1 - p(\bfx^*|E) < 1 - (1-\sum_{j=1}^n \epsilon'_j) = \sum_{j=1}^n \epsilon'_j < n\epsilon. \\
q(\bfX \neq \bfx^*|E) &= 1 - q(\bfx^*|E) < 1 - \prod_{j=1}^n (1-\epsilon'_j) \le \sum_{j=1}^n \epsilon'_j < n\epsilon.
\end{align*}
So,
\begin{equation*}
\sum_{\bfz \neq \bfx^*} |p(\bfz|E) - q(\bfz|E)| < 2n\epsilon.
\end{equation*}
Then,
\begin{align*}
\sum_{\bfz \neq \bfx^*} |p(\bfz|E) - q(\bfz|E)|^p &\le (\sup_{\bfz \neq \bfx^*} |p(\bfz|E) - q(\bfz|E)|)^{p-1} \sum_{\bfz \neq \bfx^*} |p(\bfz|E) - q(\bfz|E)| \\
&< \epsilon^{p-1} (2n\epsilon) = 2n\epsilon^p.
\end{align*}
Therefore,
\begin{equation*}
\LpDist{p}{p}{q}^p < ((n-1)\epsilon)^p + 2n\epsilon^p = ((n-1)^p+2n)\epsilon^p.
\end{equation*}
So,
\begin{equation*}
\LpDist{p}{p}{q} < ((n-1)^p+2n)^{1/p}\epsilon.
\end{equation*}
For $p=1$,
\begin{equation*}
\LpDist{1}{p}{q} < (n-1+2n)\epsilon = (3n-1)\epsilon.
\end{equation*}
And for Total Variation Distance,
\begin{equation*}
D_{TV}(p,q) = \frac{1}{2}\LpDist{1}{p}{q} < \frac{3n-1}{2}\epsilon.
\end{equation*}

\textbf{Step 4: Bound the forward KL divergence.} 

\begin{equation*}
\KLDiv{p}{q} = \sum_{\bfz} p(\bfz|E) \log \frac{p(\bfz|E)}{q(\bfz|E)} = I(X_{i_1}; \dots; X_{i_n} | E).
\end{equation*}
The conditional total correlation can be expanded using the chain rule:
\begin{equation*}
I(X_{i_1}; \dots; X_{i_n} | E) = \sum_{k=2}^n I(X_{i_k} ; X_{i_1}, \dots, X_{i_{k-1}} | E).
\end{equation*}
Each term is bounded by the conditional entropy:
\begin{equation*}
I(X_{i_k} ; X_{i_1}, \dots, X_{i_{k-1}} | E) \le H(X_{i_k} | E).
\end{equation*}
The conditional entropy $H(X_{i_k} | E)$ is bounded. Since $p_k(X_{i_k}=x_{i_k} | E) > 1-\epsilon$, it implies $p_k(X_{i_k} \neq x_{i_k} | E) = \epsilon'_k < \epsilon$.
The entropy is maximized when the remaining probability $\epsilon'_k$ is spread uniformly, leading to:
\begin{equation*}
H(X_{i_k} | E) \le H_b(\epsilon'_k) + \epsilon'_k \ln(|\calV|-1) < H_b(\epsilon) + \epsilon \ln(|\calV|-1).
\end{equation*}
Summing $(n-1)$ such terms (for $k=2, \dots, n$):
\begin{equation*}
\KLDiv{p}{q} < (n-1) [H_b(\epsilon) + \epsilon \ln(|\calV|-1)].
\end{equation*}

\end{proof}

\begin{remark}
\textbf{Assumption of a Well-Defined Joint $p_{\boldsymbol{\theta}}(X_{i_1}, \dots, X_{i_n} | E)$:}
The theorem and proof rely on $p_{\boldsymbol{\theta}}(X_{i_1}, \dots, X_{i_n} | E)$ being a well-defined joint probability mass function from which the marginals $p_{\boldsymbol{\theta}}(X_{i_j} | E)$ are consistently derived. This implies that the joint PMF is coherent and its definition does not depend on a specific factorization order beyond what is captured by the conditioning on $E$.
In practice, while MDM may not strictly satisfy this property, its behavior typically offers a close approximation.
The theorem holds for an idealized $p_{\boldsymbol{\theta}}$ that possesses these properties. As MDMs become larger and more powerful, their learned distributions might better approximate such consistency.

\textbf{Worst-Case Analysis:}
The conditions and bounds provided in the theorem (e.g., $(n+1)\epsilon \leq 1$) are derived from a worst-case analysis. This means the bounds are guaranteed to hold if the conditions are met, regardless of the specific structure of $p_{\boldsymbol{\theta}}(\bfX|E)$ beyond the high-confidence marginal property.
In practice, the actual case might be "better behaved" than the worst-case scenario. For instance, the dependencies between $X_{i_j}$ and $X_{i_k}$ (given $E$) might be weaker than what the worst-case construction assumes.
Consequently, the argmax equivalence (Result 1) might still hold frequently even if $(n+1)\epsilon$ is slightly greater than 1 (but not much larger). The condition identifies a threshold beyond which guarantees break down in the worst case, but practical performance can be more robust. Similarly, the actual $L_p$ distances or KL divergence might be smaller than the upper bounds suggest if the true joint $p_{\boldsymbol{\theta}}(\bfX|E)$ is closer to the product of marginals $q(\bfX|E)$ than the worst-case configurations.
\end{remark}

\section{Case Study}

\begin{table}[H]
\small
\caption{Qualitative comparison of responses across methods.}
\centering
\begin{tabular}{p{0.31\textwidth} p{0.31\textwidth} p{0.31\textwidth}}
\toprule
\multicolumn{3}{l}{\textbf{Prompt:} A robe takes 2 bolts of blue fiber and half that much white fiber. How many bolts in total does it take?} \\
\midrule
\textbf{Original} & \textbf{PrefixCache} & \textbf{DualCache} \\
\midrule
The robe takes 2 bolts of blue fiber.

It also takes half that much white fiber, so it takes 2/2 = 1 bolt of white fiber.

In total, the robe takes 2 + 1 = 3 bolts of fiber.
So, the value is \textcolor{red}{3} & 
The robe takes 2 bolts of blue fiber.

It also takes half that much white fiber, so it takes 2/2 = 1 bolt of white fiber.

In total, the robe takes 2 + 1 = 3 bolts of fiber.
So, the value is \textcolor{red}{3} &
The robe takes 2 bolts of blue fiber.

It also takes half that much white fiber, so it takes 2/2 = 1 bolt of white fiber.

In total, it takes 2 bolts + 1 bolt = 3 bolts of fiber.
The final result is \textcolor{red}{3} \\
\bottomrule
\end{tabular}
\label{tab:method_case}
\end{table}

\begin{table}[H]
\small
\caption{Qualitative comparison of responses with varying block size for DualCache.}
\centering
\begin{tabular}{p{0.31\textwidth} p{0.31\textwidth} p{0.31\textwidth}}
\toprule
\multicolumn{3}{l}{\textbf{Prompt:} A robe takes 2 bolts of blue fiber and half that much white fiber. How many bolts in total does it take?} \\
\midrule
\textbf{Block Size 8} & \textbf{Block Size 16} & \textbf{Block Size 32} \\
\midrule
The robe takes 2 bolts of blue fiber.

It also takes half that much white fiber, so it takes 2/2 = 1 bolt of white fiber.

In total, the robe takes 2 + 1 = 3 bolts of fiber.
So, the value is \textcolor{red}{3} & 
The robe takes 2 bolts of blue fiber.

It also takes half that much white fiber, so it takes 2/2 = 1 bolt of white fiber.

In total, the robe takes 2 + 1 = 3 bolts of fiber.
So, the value is \textcolor{red}{3} & 
The robe takes 2 bolts of blue fiber.

It also takes half that much white fiber, so it takes 2/2 = 1 bolt of white fiber.

In total, the robe takes 2 + 1 = 3 bolts of fiber.
So, the value is \textcolor{red}{3} \\
\bottomrule
\end{tabular}
\label{tab:block_case}
\end{table}
\begin{table}[H]
\small
\caption{Qualitative comparison of responses under different threshold settings.}
\centering
\begin{tabular}{p{0.31\textwidth} p{0.31\textwidth} p{0.31\textwidth}}
\toprule
\multicolumn{3}{l}{\textbf{Prompt:} A robe takes 2 bolts of blue fiber and half that much white fiber. How many bolts in total does it take?} \\
\midrule
\textbf{Threshold 0.7} & \textbf{Threshold 0.8} & \textbf{Threshold 0.9} \\
\midrule
The robe takes 2 bolts of blue fiber.

It also takes half that much white fiber, so it takes 2/2 = 1 bolt of white fiber.

In total, it takes takes 2 + 1 = 3 bolts of fiber.
So, the value is \textcolor{red}{3} \textbf{(NFE: 9)} & 
The robe takes 2 bolts of blue fiber.

It also takes half that much white fiber, so it takes 2/2 = 1 bolt of white fiber.

In total, the robe takes 2 + 1 = 3 bolts of fiber.
So, the value is \textcolor{red}{3} \textbf{(NFE: 12)} &
The robe takes 2 bolts of blue fiber.

It also takes half that much white fiber, so it takes 2/2 = 1 bolt of white fiber.

In total, the robe takes 2 + 1 = 3 bolts of fiber.
So, the value is \textcolor{red}{3} \textbf{(NFE: 20)} \\
\bottomrule
\end{tabular}
\label{tab:threshold_case}
\end{table}
\subsection{Effect of Caching Strategies on Response Quality}

Table~\ref{tab:method_case} qualitatively compares answers from the Original, PrefixCache, and DualCache methods for the arithmetic prompt. All correctly compute the answer (3 bolts), following similar step-by-step reasoning, with only minor differences in phrasing. This shows cache strategies maintain answer accuracy and logical clarity while improving efficiency; semantic fidelity and interpretability are unaffected.

\subsection{Effect of Block Size in DualCache}

Table~\ref{tab:block_case} examines different block sizes (8, 16, 32) in DualCache. For this arithmetic prompt, all settings yield correct, clearly explained answers with no meaningful output differences. Thus, DualCache is robust to block size for such problems, allowing efficiency improvements without compromising quality.

\subsection{Impact of Dynamic Threshold Settings}

Table~\ref{tab:threshold_case} investigates dynamic threshold values (0.7, 0.8, 0.9). The model consistently produces the correct answer and clear explanations, regardless of threshold. While higher thresholds increase computational effort (NFE from 9 to 20), answer quality remains stable, indicating threshold adjustment mainly affects efficiency, not correctness, for straightforward arithmetic questions.

\subsection{Multimodal Generation with \texttt{LLAda-V}}
\label{subsec:llada-v-case-study}

\begin{figure}[ht]
    \centering
    \includegraphics[width=\linewidth]{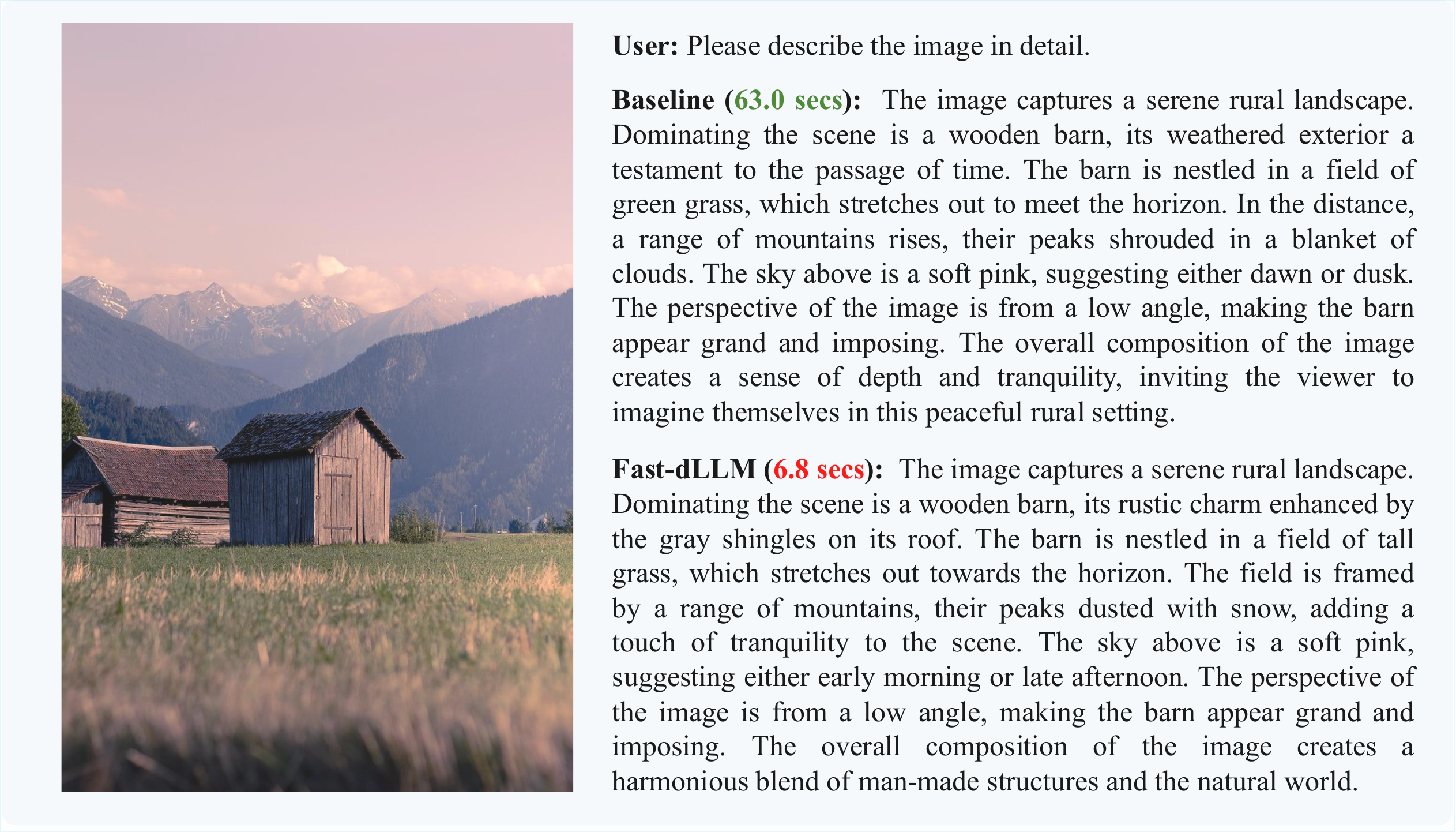
    }
    \caption{Comparison between the baseline and Fast-dLLM on a visual description task. Fast-dLLM produces a comparable and faithful image caption in a fraction of the decoding time.}
    \label{fig:llada-v-case}
\end{figure}

To qualitatively analyze the effectiveness of our Fast-dLLM framework in multimodal scenarios, we conduct a visual case study where the model is tasked with generating a highly detailed image description. As illustrated in Figure~\ref{fig:llada-v-case}, both the baseline model and our Fast-dLLM are given the same visual input and user prompt: ``\textit{Please describe the image in detail}.''

The baseline model requires 63.0 seconds to complete the generation, producing a detailed and poetic description of the rural landscape. It highlights elements such as the weathered wooden barn, the soft pink sky, and the tranquil atmosphere.

In contrast, our Fast-dLLM completes the task in just 6.8 seconds---a nearly 10$\times$ speedup---while maintaining rich visual detail. It further enhances the description with additional grounding (e.g., ``gray shingles on its roof'', ``touch of tranquility’’), reflecting a strong alignment with both appearance and mood cues from the image. Notably, the generated caption retains compositional depth and stylistic fluency, illustrating the model's ability to balance fluency and factuality even under diffusion-based parallel decoding.

This case highlights how LLAda-V with Fast-dLLM decoding enables high-quality vision-language generation at significantly improved efficiency, paving the way for faster and more interactive multimodal applications.

\section{Experiment Details}
\subsection{Further Experiments with LLaDA-V}

\begin{table}[h!]
\centering
\caption{Effect of block length on performance (MathVista, 48 Steps)}
\begin{tabular}{lccccc}
\toprule
\textbf{Block Length} & 4 & 8 & 16 & 32 & 96 \\
\midrule
Accuracy (\%) & 51.2 & 50.7 & 51.8 & 52.3 & 59.7 \\
Throughput (tok./s) & 6.1 & 6.2 & 5.5 & 5.5 & 5.6 \\
\bottomrule
\end{tabular}
\label{tab:blocklen_effect}
\end{table}

\begin{table}[h!]
\centering
\caption{MathVista Performance with \method at different refresh intervals (block length = 96)}
\begin{tabular}{lccccc}
\toprule
\textbf{Refresh Interval} & 2 & 4 & 8 & 16 & 32 \\
\midrule
Accuracy (\%) & 59.2 & 59.2 & 58.2 & 57.1 & 56.6 \\
Throughput (tok./s) & 15.9 & 19.5 & 21.1 & 25.2 & 28.2 \\
\bottomrule
\end{tabular}
\label{tab:refresh_effect}
\end{table}

In Table~\ref{tab:blocklen_effect}, we investigate how the choice of block length affects the performance of LLaDA-V on MathVista under a fixed decoding length of 48 steps. The results show that the model achieves the highest accuracy with a block length of 96. However, when reducing the block size to 8 or 4, the accuracy drops significantly by over 8\%.

Given this sensitivity to block length, we choose not to break the output into small blocks for updating caches individually. Instead, we keep the block length fixed at 96 and adopt a refresh-based strategy: the cache is updated only every $r$ decoding steps using the most recent full block. As shown in Table~\ref{tab:refresh_effect}, increasing the refresh interval leads to consistent gains in throughput—from 15.9 tokens/s at interval 2 to 28.2 tokens/s at interval 32. While accuracy drops slightly with larger intervals, it remains above 56.6\%, suggesting that aggressive refresh scheduling can yield substantial speedups with only minor performance degradation.

\subsection{Performance Comparison between Threshold and Factor Strategy}

\begin{table}[h]
\centering
\renewcommand{\arraystretch}{1.3}
\setlength{\tabcolsep}{14pt}
\caption{
Performance comparison between \textbf{Threshold} and \textbf{Factor} confidence-aware decoding on GSM8K and MATH benchmarks with generation lengths of 256 and 512. Each block shows accuracy (top row) and throughput with speedup (bottom row). Factor decoding provides favorable trade-offs in most settings.
}
\begin{tabular}{l|c|cc}
\toprule
\textbf{Benchmark} & \textbf{Gen. Len} & \textbf{Threshold} & \textbf{Factor} \\
\midrule
\multirow{2}{*}{} 
    & 256 
      & 78.5 
      & 77.5 \\
    GSM8K (5-shot) &     
      & \textcolor{blue}{54.4} (\textcolor{orange}{8.1×}) 
      & \textcolor{blue}{78.5} (\textcolor{orange}{11.7x}) \\
\cline{2-4}
\multirow{2}{*}{} 
    & 512 
      & 77.2 
      & 74.8 \\
    &     
      & \textcolor{blue}{35.3} (\textcolor{orange}{11.0×}) 
      & \textcolor{blue}{47.1} (\textcolor{orange}{14.7x}) \\
\midrule
\multirow{2}{*}{} 
    & 256 
      & 33.2 
      & 32.0 \\
    MATH (4-shot)&     
      & \textcolor{blue}{51.7} (\textcolor{orange}{5.7×}) 
      & \textcolor{blue}{78.3} (\textcolor{orange}{8.6x}) \\
\cline{2-4}
\multirow{2}{*}{} 
    & 512 
      & 36.0 
      & 35.2 \\
    &     
      & \textcolor{blue}{47.1} (\textcolor{orange}{5.9×}) 
      & \textcolor{blue}{64.6} (\textcolor{orange}{8.1x}) \\
\bottomrule
\end{tabular}

\label{tab:threshold-vs-factor}
\end{table}

We compare the performance of our threshold-based and factor-based confidence-aware parallel decoding strategies on GSM8K and MATH benchmarks (Table~\ref{tab:threshold-vs-factor}). While the threshold strategy achieves marginally better accuracy in most settings (e.g., 78.5\% vs. 77.5\% on GSM8K with 256 tokens), the factor strategy demonstrates substantially superior throughput performance. 

Specifically, factor decoding achieves 1.4-1.5× higher throughput than threshold decoding across all settings. On GSM8K with 256 tokens, factor decoding reaches 78.5 tokens/sec (11.7× speedup) compared to 54.4 tokens/sec (8.1× speedup) for threshold decoding. This throughput advantage becomes even more pronounced on longer generation tasks—for GSM8K with 512 tokens, factor decoding attains 47.1 tokens/sec while threshold only achieves 35.3 tokens/sec.

The results demonstrate that factor decoding offers a compelling trade-off: it sacrifices minimal accuracy (typically 1-3\%) in exchange for significant throughput improvements (40-50\% higher). This makes factor decoding particularly attractive for latency-sensitive applications where the slight accuracy reduction is acceptable. The consistent pattern across both benchmarks and generation lengths validates the robustness of the factor strategy's theoretical foundation, which adaptively controls parallelism based on the confidence bound $(n+1)\epsilon < f$.

\begin{figure}[h]
    \centering
    \includegraphics[width=\linewidth]{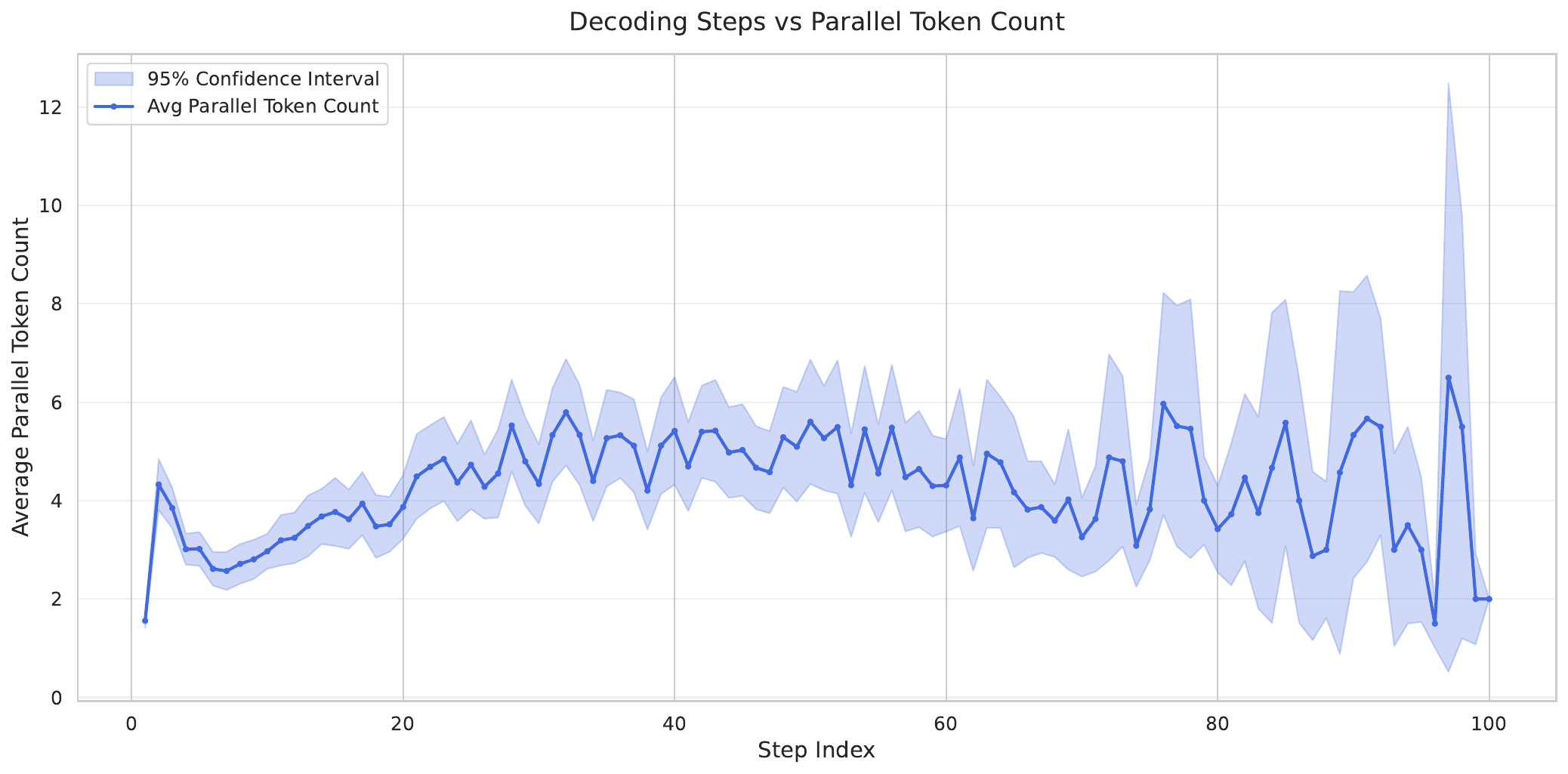}
    \caption{Average number of tokens generated at each decoding step. Blue line shows the mean token count, and the shaded area denotes the 95\% confidence interval.}
    \label{fig:avg-token-count}
\end{figure}

\subsection{Comparison between LLaDA and LLaDA-1.5}

We compare the performance of LLaDA and its enhanced version LLaDA-1.5 across both GSM8K (5-shot) and MATH (4-shot) benchmarks under two generation length settings (256 and 512 tokens), as shown in Table~\ref{tab:llada_1.5}. Each cell reports accuracy and decoding throughput (in tokens per second), along with the relative speedup over the greedy baseline.

Across GSM8K settings, LLaDA-1.5 consistently improves accuracy over the original LLaDA, achieving a notable +2.2\% absolute gain at 256-token generation and +3.2\% at 512-token generation. Furthermore, it maintains strong decoding efficiency, with throughput reaching 59.4 tokens/sec at 256 tokens, improving upon LLaDA’s 54.1 tokens/sec under the same setting.

On the MATH benchmark, accuracy between the two versions remains comparable. However, LLaDA-1.5 slightly improves throughput at 256 tokens (53.7 vs. 51.7) while incurring a mild efficiency regression at the 512-token setting (41.1 vs. 47.1). This suggests that while LLaDA-1.5 introduces enhancements beneficial for shorter or moderate decoding contexts, longer sequences may require further optimization.

Overall, LLaDA-1.5 consistently provides either superior accuracy or better decoding speed across settings, demonstrating better performance-efficiency trade-offs and highlighting the benefit of incorporating adaptive improvements on top of the base LLaDA architecture.

\begin{table*}[t!]
\centering
\caption{
Performance comparison between LLaDA and LLaDA-1.5. Each cell presents the accuracy and the decoding throughput in tokens per second with relative speedup to the LLaDA baseline (bottom row, \textcolor{blue}{blue: tokens per second}/\textcolor{orange}{orange: relative speedup}).}
\begin{tabular}{lc!{\vrule width 1pt}cc}
\toprule
Benchmark & Gen Length & LLaDA (\method) & LLaDA 1.5 (\method) \\
\midrule
\multirow{2}{*}[-1.8ex]{\centering GSM8K (5-shot)}
    & 256 & 78.5 & 80.7 \\
    &       & \textcolor{blue}{54.1} (\textcolor{orange}{8.1$\times$}) 
            & \textcolor{blue}{59.4} (\textcolor{orange}{8.9$\times$}) \\
    & 512 & 77.2 & 80.4 \\
    &       & \textcolor{blue}{35.3} (\textcolor{orange}{11.0$\times$}) 
            & \textcolor{blue}{33.0} (\textcolor{orange}{10.3$\times$}) \\
\midrule
\multirow{2}{*}[-1.8ex]{\centering MATH (4-shot)}
    & 256 & 33.2 & 32.6 \\
    &       & \textcolor{blue}{51.7} (\textcolor{orange}{5.7$\times$})   
            & \textcolor{blue}{53.7} (\textcolor{orange}{5.9$\times$})\\
    & 512 & 36.0 & 35.1 \\
    &       & \textcolor{blue}{47.1} (\textcolor{orange}{5.9$\times$})
            & \textcolor{blue}{41.1} (\textcolor{orange}{5.1$\times$})\\
\bottomrule
\label{tab:llada_1.5}
\end{tabular}
\vspace{-10pt}
\end{table*}

\subsection{Analysis of Parallel Token Counts across Decoding Steps}
\label{subsec:token-count-stats}

\begin{figure*}[h]
    \centering
    \includegraphics[width=\textwidth]{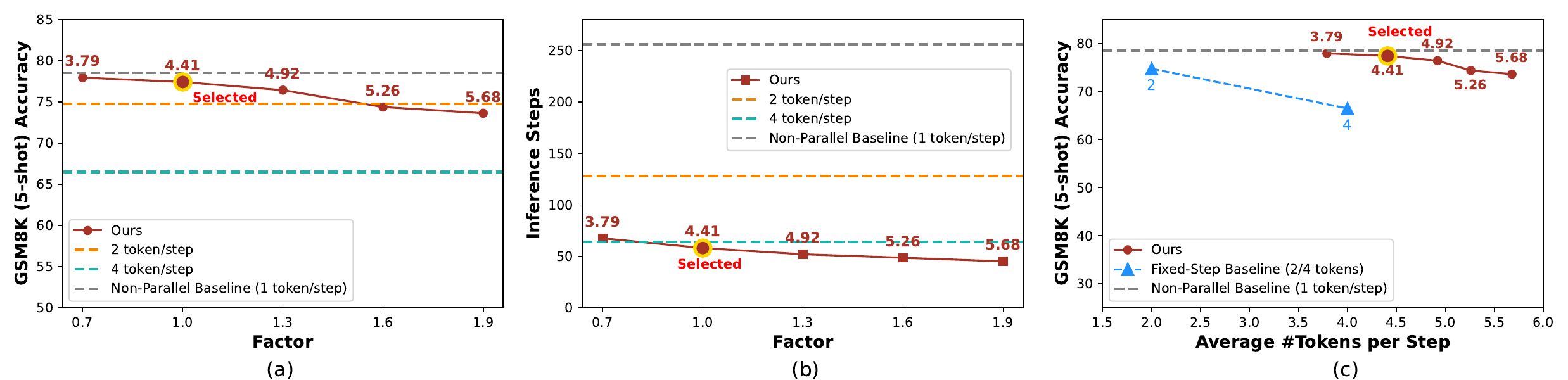} 
    \caption{
(a) GSM8K (5-shot) accuracy across different factor values using our factor-based decoding strategy. Numbers above each point indicate the average number of tokens decoded per step. The dashed lines show the accuracy of the baseline method with 2 or 4 tokens per step, and the non-parallel (1 token/step) baseline.
(b) The corresponding number of inference steps needed under each factor setting. Our method generally requires significantly fewer steps than fixed-step baselines.
(c) Accuracy versus average number of tokens decoded per step on GSM8K (5-shot). Our factor-based decoding achieves better accuracy-efficiency trade-offs compared to baselines. The red “Selected” point represents the setting chosen in our main results.
}
    \label{fig:factor-metrics}
\end{figure*}

To better understand the behavior of factor-based parallel generation, we analyze the average number of tokens generated at each decoding step. Specifically, we collect statistics from all intermediate steps of the sampling process and compute the average number of tokens generated in parallel per step. The results are visualized in Figure~\ref{fig:avg-token-count}, along with a 95\% confidence interval indicating cross-sample variability.

As shown in Figure~\ref{fig:avg-token-count}, the average number of tokens generated in parallel gradually increases during the early to middle stages of decoding, peaking roughly between step 30 to step 60. After this peak, the parallelism tends to slightly decline toward the end of generation. This suggests that the model becomes more confident in generating outputs during the mid-decoding phase, allowing it to produce more tokens simultaneously. Toward the final steps, the decoding process tends to become more conservative, reducing the number of tokens produced at each step.

The shaded confidence interval reveals greater variance in later decoding steps, indicating instability and inconsistent generation behavior across samples. This is expected since tail-end decoding steps tend to handle only a few remaining tokens required to complete the output, and the number of remaining tokens could differ widely among different samples (e.g., due to early completion or padding).

These observations are important for understanding how decoding efficiency can be optimized: increasing parallelism during high-confidence phases (middle steps) offers computational savings, while conservative behavior near boundaries maintains quality.

\subsection{Throughput Comparison under Varying Batch Sizes}
\label{subsection:throughput-comparison}
\begin{figure}[t!]
    \centering
    \includegraphics[width=0.9\linewidth]{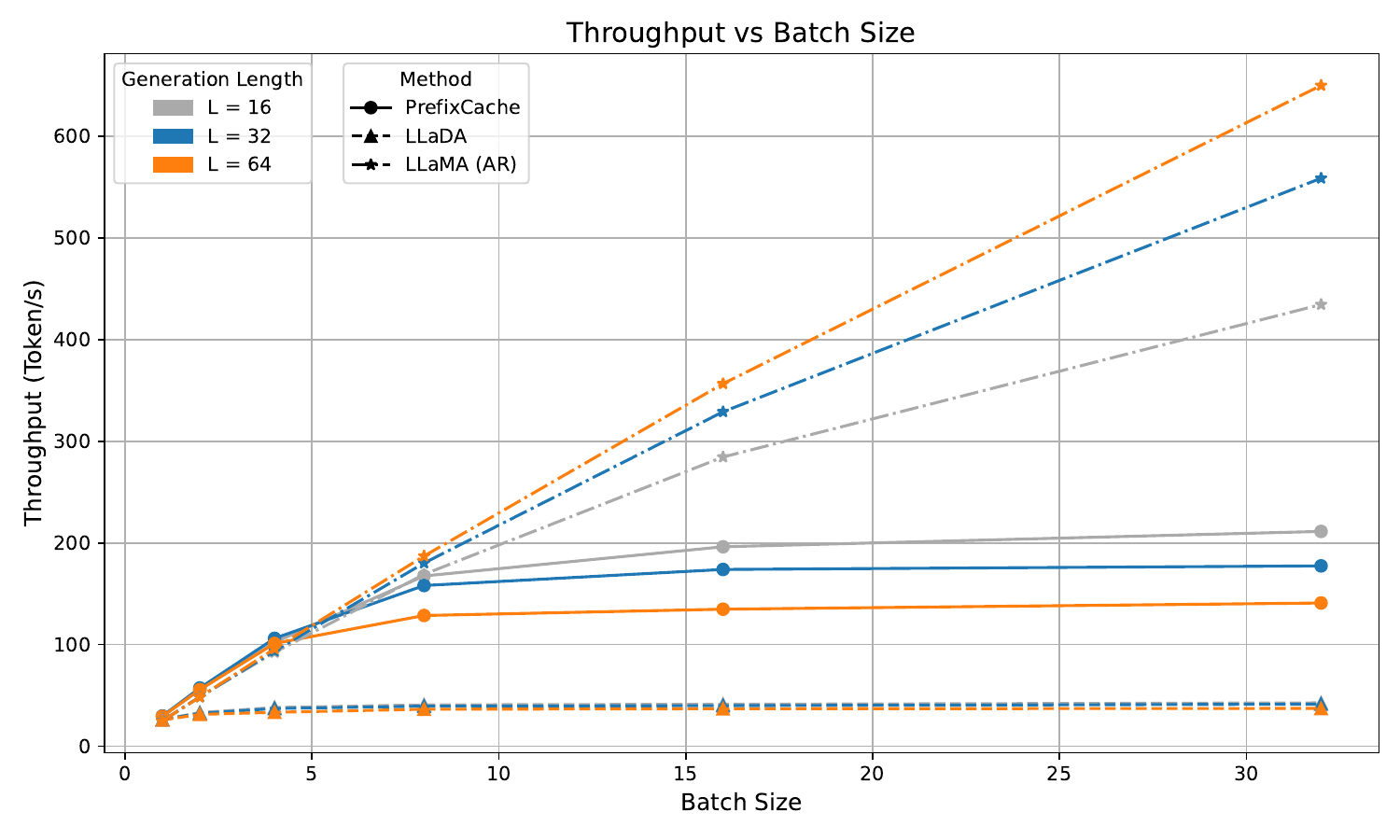}
    \caption{
        Throughput comparison between \textbf{PrefixCache}, \textbf{LLaDA}, and \textbf{LLaMA} under different generation lengths and batch sizes. 
        All models are evaluated on an \textbf{NVIDIA A100} GPU with the prefill length fixed at 256.
    }
    \label{fig:throughput_varying_batch}
\end{figure}

All experiments are conducted on an NVIDIA A100 GPU, with the prefill length fixed to 256 tokens. The generation length is varied among 16, 32, and 64 tokens, and batch sizes range from 1 to 32. This setup reflects realistic deployment scenarios, allowing the evaluation of decoding efficiency under diverse conditions. 

It should be noted that parallel decoding allows multiple tokens to be generated simultaneously affected by dummy input tokens. To ensure fairness, we focus solely on the acceleration provided by caching techniques.

PrefixCache is designed as an acceleration mechanism for LLaDA, a diffusion-based LLM, and successfully boosts the throughput significantly. Figure~\ref{fig:throughput_varying_batch} shows that \textbf{PrefixCache achieves consistent improvements across all batch sizes and generation lengths}, making it particularly suited for scenarios with smaller generation lengths and larger batch sizes. For instance, with a generation length of 16 and batch size of 32, PrefixCache achieves a throughput of over 211 tokens/s, significantly outperforming the native LLaDA which reaches only 43 tokens/s, demonstrating nearly $5\times$ improvement.

While LLaDA exhibits limited scalability with increasing batch sizes—its throughput plateaus after batch size 8—this limitation is inherent to diffusion-based LLMs, which are compute-bound by nature. In contrast, LLaMA, an autoregressive (AR) model, benefits greatly from large batch sizes. As the batch size increases, LLaMA shifts from being memory-bound to compute-bound, allowing it to achieve high absolute throughput at larger batch settings.

These results highlight the practical advantages of PrefixCache in accelerating compute-bound diffusion models like LLaDA, especially for latency-critical and high-throughput applications. Furthermore, the scalability and efficiency provided by PrefixCache bridge the gap between diffusion-based LLMs and AR models like LLaMA, showcasing its importance for large-scale deployment settings.

\clearpage

{
    \small
    \bibliographystyle{plain}
    \bibliography{main}
}

\end{document}